%% file: main.tex
\documentclass[sigconf]{acmart}
\acmSubmissionID{267}

\usepackage{booktabs} %
\usepackage{multirow}
\usepackage{xcolor}
\usepackage{soul}
\usepackage{enumitem}
\usepackage{url}
\usepackage[ruled]{algorithm2e} %

\SetAlFnt{\small}
\SetAlCapFnt{\small}
\SetAlCapNameFnt{\small}
\SetAlCapHSkip{0pt}

\copyrightyear{2023}
\acmYear{2023}
\setcopyright{rightsretained}
\acmConference[SIGGRAPH '23 Conference Proceedings]{Special Interest Group on Computer Graphics and Interactive Techniques Conference Conference Proceedings}{August 6--10, 2023}{Los Angeles, CA, USA}
\acmBooktitle{Special Interest Group on Computer Graphics and Interactive Techniques Conference Conference Proceedings (SIGGRAPH '23 Conference Proceedings), August 6--10, 2023, Los Angeles, CA, USA}
\acmDOI{10.1145/3588432.3591490}
\acmISBN{979-8-4007-0159-7/23/08}

\begin{document}
\title{PoseVocab: Learning Joint-structured Pose Embeddings for Human Avatar Modeling}

\author{Zhe Li}
\orcid{0000-0003-4703-0875}
\affiliation{%
 \institution{Tsinghua University}
 \city{Beijing}
 \country{China}}
\email{liz19@mails.tsinghua.edu.cn}

\author{Zerong Zheng}
\orcid{0000-0003-1339-2480}
\affiliation{%
 \institution{Tsinghua University \& NNKosmos Technology}
 \city{Beijing}
 \country{China}}
\email{zzr18@mails.tsinghua.edu.cn}

\author{Yuxiao Liu}
\orcid{0009-0002-0580-6647}
\affiliation{%
 \institution{Tsinghua University}
 \city{Shenzhen}
 \country{China}}
\email{liuyuxia22@mails.tsinghua.edu.cn}

\author{Boyao Zhou}
\orcid{0009-0004-4583-2676}
\affiliation{%
 \institution{Tsinghua University}
 \city{Beijing}
 \country{China}}
\email{bzhou22@mail.tsinghua.edu.cn}

\author{Yebin Liu}
\orcid{0000-0003-3215-0225}
\affiliation{%
 \institution{Tsinghua University}
 \city{Beijing}
 \country{China}}
 \email{liuyebin@mail.tsinghua.edu.cn}

\renewcommand{\shortauthors}{Li, Z. et al}

\input{secs/0_abstract.tex}




\begin{teaserfigure}
\centering
\includegraphics[width=\linewidth]{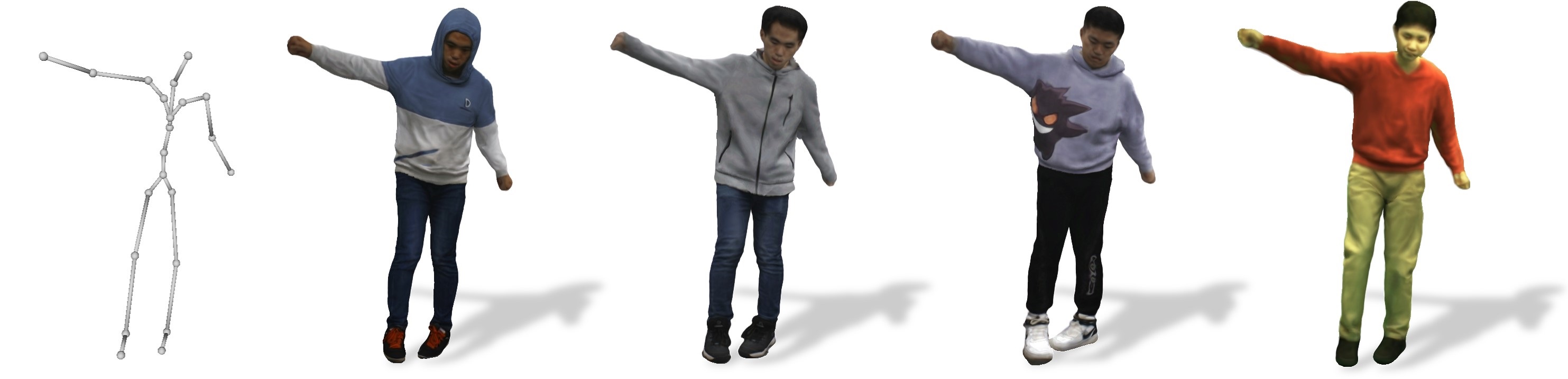}
\caption{Our method can create animatable avatars with realistic pose-dependent details from multi-view RGB videos.}
\Description{Our method can create animatable avatars with realistic pose-dependent details from multi-view RGB videos.}
\label{fig:teaser}
\end{teaserfigure}

\maketitle

\input{secs/1_introduction}
\input{secs/2_related_work}
\input{secs/3_method}

\input{secs/4_results}

\input{secs/5_discussion}
\input{secs/7_supp}

\bibliographystyle{ACM-Reference-Format}
\bibliography{ref}

\input{secs/6_figure_only}

\end{document}

%% file: secs/0_abstract.tex
\begin{abstract}
     Creating pose-driven human avatars is about modeling the mapping from the low-frequency driving pose to high-frequency dynamic human appearances, so an effective pose encoding method that can encode high-fidelity human details is essential to human avatar modeling.
     To this end, we present PoseVocab, a novel pose encoding method that encourages the network to discover the optimal pose embeddings for learning the dynamic human appearance.
     Given multi-view RGB videos of a character, PoseVocab constructs key poses and latent embeddings based on the training poses.
     To achieve pose generalization and temporal consistency, we sample key rotations in $so(3)$ of each joint rather than the global pose vectors, and assign a pose embedding to each sampled key rotation.
     These joint-structured pose embeddings not only encode the dynamic appearances under different key poses, but also factorize the global pose embedding into joint-structured ones to better learn the appearance variation related to the motion of each joint.
     To improve the representation ability of the pose embedding while maintaining memory efficiency, we introduce feature lines, a compact yet effective 3D representation, to model more fine-grained details of human appearances.
     Furthermore, given a query pose and a spatial position, a hierarchical query strategy is introduced to interpolate pose embeddings and acquire the conditional pose feature for dynamic human synthesis.
     Overall, PoseVocab effectively encodes the dynamic details of human appearance and enables realistic and generalized animation under novel poses.
     Experiments show that our method outperforms other state-of-the-art baselines both qualitatively and quantitatively in terms of synthesis quality.
     Code is available at \url{https://github.com/lizhe00/PoseVocab}.
\end{abstract}

%% file: secs/1_introduction.tex
\section{Introduction}

Human avatar modeling, due to its potential value in holographic conferences, Metaverse, game and movie industries, has been a popular topic in computer graphics and vision for decades.
Animatable human avatars usually take the skeletal pose as the driving signal and output realistic human models with pose-dependent dynamic details.
Many human avatar techniques \cite{bagautdinov2021driving,peng2021animatable,liu2021neural,zheng2022structured} utilize neural networks to model the mapping from the pose input to the dynamic human appearance.
However, how to effectively encode the pose input into the network still remains a challenging problem.

Many previous works take SMPL-derived \cite{loper2015smpl} attributes like pose vectors \cite{zheng2022structured,li2022tava,saito2021scanimate} or SMPL positional maps \cite{ma2021power} as the conditional pose features. 
Then a neural network, usually an MLP, is trained to map 3D positions and the corresponding pose features to a 3D representation, e.g., mesh, point cloud, signed distance field (SDF), and radiance field (NeRF) \cite{mildenhall2020nerf}, to model the dynamic human appearance.
{Unfortunately, the change of the driving signal, i.e., SMPL-derived attributes, is low-frequency, while the human appearance varies at a much higher frequency. Therefore, it is challenging for MLPs to model the mapping due to the low-frequency bias of MLPs \cite{tancik2020fourier}, thus yielding blurry appearances without fine-grained garment wrinkles.}

{Previous works \cite{instantngp,yu2021plenoctrees,Chen2022TensoRF} demonstrate that learnable latent embeddings at NeRF input end can encode much more high-frequency details for static scene rendering. To extend such embeddings to dynamic human modeling, inspired by word embeddings in word2vec \cite{mikolov2013efficient,mikolov2013distributed}, we propose PoseVocab that consists of pairs of key poses and learnable pose embeddings, to encourage the network to discover the optimal embeddings for encoding high-frequency human appearances under various poses.}
However, naively constructing pairs of global poses and global latent codes like \cite{peng2021animatable} yields poor generalization to unseen poses because the global pose vector entangles the information from all the joints.
What's worse, these global codes fail to encode fine-grained details of human appearances due to their limited representation capacity. 
Therefore, we propose \textit{joint-structured pose embeddings} by sampling key rotations in $so(3)$ of each joint and assigning them the corresponding latent embeddings. These joint-structured pose embeddings are distributed in the rotation domain of each joint, and serve as the discrete samples of the continuous pose feature space for interpolation.
Furthermore, to guarantee both spatial capacity and memory efficiency, each pose embedding is defined as three \textit{feature lines} along $x$, $y$ and $z$ axes.
Similar to EG3D \cite{chan2022efficient} and TensoRF \cite{Chen2022TensoRF}, the feature lines decompose a 3D volume into three axes via orthogonal projections.
Compared with a global latent code, feature lines demonstrate a more powerful capacity for encoding fine-grained dynamic details while maintaining memory efficiency.

Based on the constructed PoseVocab, a \textit{hierarchical query} strategy is introduced for avatar animation. 
Our hierarchical query includes three levels: the joint level, the key rotation level and and the spatial level.
Specifically, given a query body pose, we first decompose the global pose vector into rotations of each joint.
For each query joint rotation, we search for K nearest neighbors (KNN) key rotations and interpolate the corresponding pose embeddings, i.e., feature lines.
In the spatial level, the pose feature of a query 3D position is sampled by orthogonal projections on the interpolated feature lines.
Finally, the 3D position and the corresponding pose feature are fed into an MLP to decode a neural radiance field (NeRF) \cite{mildenhall2020nerf} that represents the dynamic 3D character.
The hierarchical query in PoseVocab not only decomposes the effects of each joint rotation on the dynamic appearance for better generalization to novel poses, but also guarantees the temporally consistent animation benefiting from the smooth KNN interpolation.

In summary, our technical contributions are below:

\begin{itemize}[leftmargin=*]
    \item Joint-structured pose embeddings that not only disentangle the effects of different joints on the dynamic appearance, but also encode high-frequency details for realistic avatar modeling. %
    \item Feature lines, a new compact yet effective 3D representation that improves the representation ability of pose embeddings while maintaining memory efficiency. %
    \item A hierarchical query strategy in PoseVocab that interpolates joint-structured pose embeddings in the joint, key-rotation and spatial levels for generalized and temporally consistent animation. %
\end{itemize}

Compared with other pose encoding methods, PoseVocab not only has the ability to encode the high-frequency human dynamic appearance, but also generalizes well to novel poses. 
Overall, our method can automatically create a realistic animatable avatar represented by a pose-conditioned NeRF from multi-view videos,
and experiments show that our method outperforms other state-of-the-art approaches both qualitatively and quantitatively.

%% file: secs/2_related_work.tex
\section{Related Work}

Human avatar modeling is a popular research topic in the last few years, and many methods have been proposed to reconstruct animation-ready avatars from short videos \cite{alldieck2018video,alldieck2019tex2shape,jiang2022selfrecon,jiang2022neuman,te2022neural,peng2022selfnerf,jiang2022instantavatar,su2021a-nerf,Feng2022scarf} or single images \cite{huang2020arch,he2021arch++,huang2022one}. Unfortunately, they cannot synthesize pose-dependent appearances like dynamic cloth wrinkles. In this paper, we aim to model these pose-dependent appearance details, so we mainly review the related methods that are able to achieve similar goals.

\subsection{Pose Encoding in Avatar Modeling}

{Human pose encoding is widely explored, and many representations including pose vectors, 6D representations \cite{zhou2019continuity} and latent codes \cite{SMPL-X:2019} are proposed for pose priors \cite{SMPL-X:2019,tiwari2022pose} and motion generation \cite{guo2022generating,tevet2022motionclip}. How to effectively encode the pose input into the network is also one of the core problems in avatar modeling.}
Many works take SMPL-derived \cite{loper2015smpl} attributes to encode the pose input.
In the geometric avatar modeling, given 3D scans or depth sequences of a character, SCANimate \cite{saito2021scanimate}, SNARF \cite{chen2021snarf}, Neural-GIF \cite{tiwari2021neural}, NASA \cite{deng2020nasa}, LEAP \cite{mihajlovic2021leap}, DSFN \cite{burov2021dynamic}, PINA \cite{dong2022pina} and LaplacianFusion \cite{kim2022laplacianfusion} adopted SMPL pose vectors or joint rotations as the pose condition to learn the pose-dependent implicit surfaces or displacement fields.
MetaAvatar \cite{wang2021metaavatar} proposed to learn an avatar from only a few depth images with the meta-learned network as the initialization. 
SCALE \cite{ma2021scale}, GeoTexAvatar \cite{li2022avatarcap}, FITE \cite{lin2022learning} and CLoSET \cite{zhang2023closet} took UV or rendered positional maps of posed SMPL models as the pose condition to regress the pose-dependent warping field. 
{COAP \cite{mihajlovic2022coap} proposed a part-aware neural network to condition the body shape on local SMPL point clouds.}

On the other hand, given RGB videos of a character under various poses, many textured avatar modeling methods also adopted similar SMPL-derived attributes as the pose feature to represent the dynamic human appearance.
Specifically, \cite{zheng2022structured,zheng2023avatar} defined a set of local radiance fields attached to sampled SMPL nodes, and learned the mapping from SMPL pose vectors to the node residual and varying details of the human appearance.
TAVA \cite{li2022tava} proposed to jointly model the non-rigid warping field and shading effects directly conditioned on the pose vectors.
{Besides, pose vector-based encoding is also adopted in animatable hand modeling \cite{corona2022lisa}.}
{\cite{yoon2022learning} utilized SMPL normal maps and velocities as pose conditions and took SMPL as a 3D proxy for deferred neural rendering \cite{thies2019deferred}.}
{DANBO \cite{su2022danbo} regressed human appearances from 6D representations \cite{zhou2019continuity} of skeletal poses via GNN \cite{kipf2016semi}.}
However, the above low-frequency SMPL-derived attributes have a poor capacity to represent the high-fidelity dynamic human appearance, thus essentially limiting the synthesis quality of avatars.
{Neural Actor \cite{liu2021neural} regressed texture maps from normal maps by image-to-image translation with adversarial loss in SMPL UV space, but such an encoding strategy constrains the performer to wear tight clothes for topological consistency with SMPL.}
On the other end of the spectrum, Neural Body \cite{peng2021neural}, Animatble NeRF \cite{peng2021animatable,peng2022animatable}, ARAH \cite{wang2022arah} and TotalSelfScan \cite{dong2022totalselfscan} assigned a global latent code for each training frame to compensate for the time-varying dynamic appearance and learned these codes in an auto-decoding fashion \cite{park2019deepsdf}.
Although these codes ``offload'' the network representation power to themselves to encode varying details, they fail to encode fine-grained appearance details due to their limited representation capacity. Furthermore, global latent codes entangle the information from all the body joints, exacerbating their generalization capability to novel poses.

\subsection{Avatars with Reconstruction, Tracking, or Simulation}
Many works aimed to combine hybrid technologies like dense reconstruction, non-rigid tracking, and cloth simulation for creating more realistic avatars from multi-view videos.
\cite{bagautdinov2021driving}, \cite{xiang2021modeling} and \cite{halimi2022pattern} first reconstructed and tracked 3D meshes of the whole body or garments, and then learned the deformation and varying appearances with the dense correspondences as a strong prior.
\cite{xu2011video} and \cite{habermann2021real} relied on a template mesh for blending textures from the database or learning the deformation and textures, respectively.
Dressing Avatars \cite{xiang2022dressing} utilized high-fidelity tracking of dense geometry in the training stage, whereas in the animation this method applied a cloth simulator to generate more realistic cloth dynamics.
\cite{remelli2022drivable} extended the driving signal from skeletal poses to sparse images, and leveraged texel-aligned features to synthesize more realistic details.
Compared with these hybrid methods, our method does not rely on any preprocessing steps like reconstruction, tracking or simulation, and can be trained in an end-to-end manner given the RGB videos and SMPL registrations.

%% file: secs/3_method.tex
\begin{figure*}[t]
    \centering
    \includegraphics[width=\linewidth]{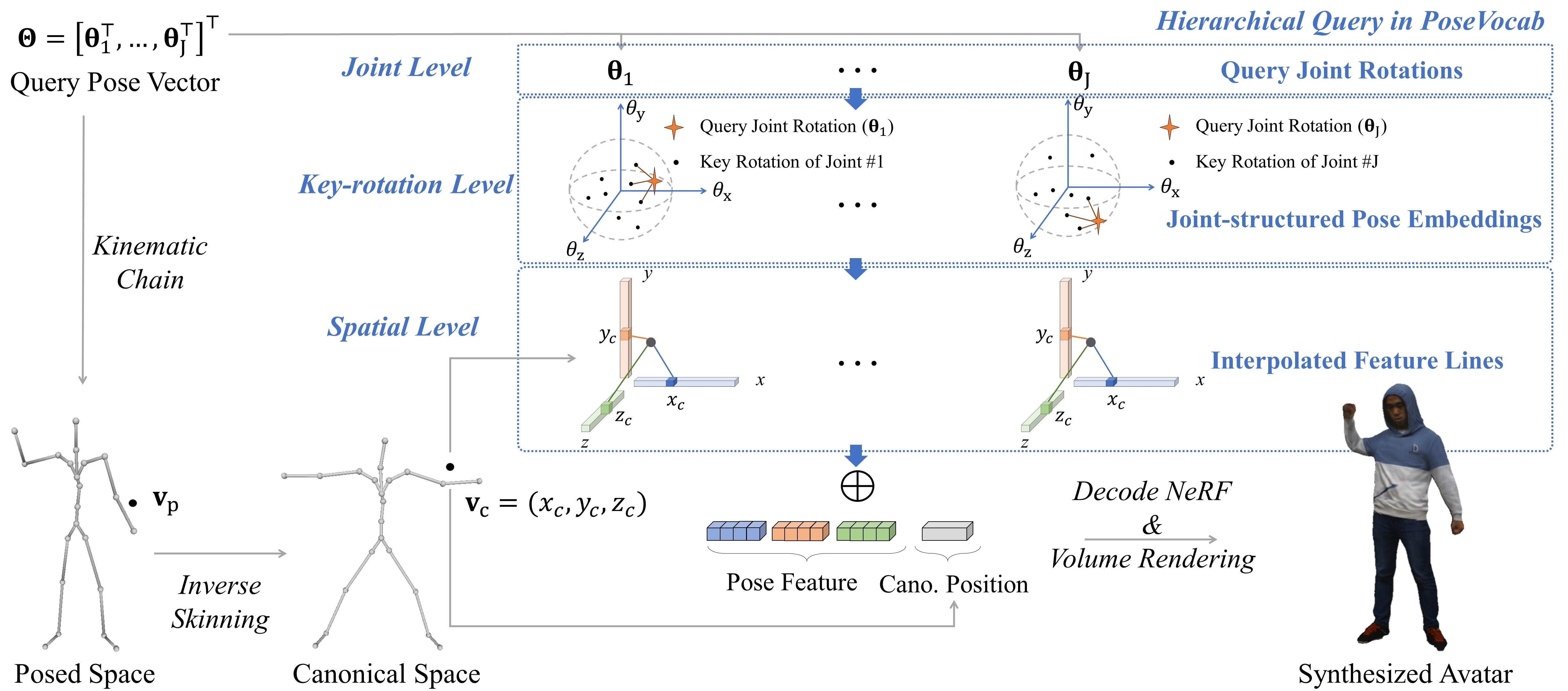}
    \caption{\textbf{Overview of the representation of PoseVocab.} 
    PoseVocab is constructed based on the training poses by sampling the key rotations in $so(3)$ of each joint and assigning a pose embedding for each key rotation.
    These joint-structured pose embeddings encode the dynamic appearance of the character under various poses.
    Given a query pose and 3D position, we hierarchically interpolate pose embeddings in joint, key-rotation and spatial levels to acquire the conditional pose feature, which is fed into an MLP to decode the radiance field, eventually synthesizing the high-fidelity human avatar via volume rendering.
    }
    \label{fig:overview}
\end{figure*}

\section{Method}

Given multi-view RGB videos of a character with $T$ frames captured by $N$ cameras, we denote the RGB sequences as $\{\mathbf{I}_n^t|\,1\leq n\leq N,\,1\leq t\leq T\}$.
We assume that the body pose of each frame is known, and denoted as $\mathbf{\Theta}^t\in \mathbb{R}^{3\times J}$, where $J$ is the joint number of the human body.
The human avatar modeling problem is to model the mapping from the body pose $\mathbf{\Theta}^t$ to the dynamic human appearance described by the corresponding RGB images.

Similar to other avatar representations \cite{peng2021animatable,liu2021neural}, we first factor out the rigid skeletal motions by deforming the 3D human from the posed space to the canonical one using linear blend skinning (LBS).
Then we represent the canonical 3D human as a pose-conditioned neural radiance field (NeRF) \cite{mildenhall2020nerf} that takes a canonical 3D position {$\mathbf{v}_c$, a view direction $\mathbf{d}$ and the pose feature $\mathbf{f}(\mathbf{\Theta}^t,\mathbf{v}_c)$} as input and returns a tuple of color $\mathbf{c}$ and SDF $s$, i.e.,
\begin{equation}
\label{eq:cano nerf}
    g:(\mathbf{v}_c, {\mathbf{d}},\mathbf{f}(\mathbf{\Theta}^t,\mathbf{v}_c))\rightarrow (\mathbf{c},s).
\end{equation}
{In our method, we follow VolSDF \cite{yariv2021volume} to convert SDF to density for SDF-based volume rendering.}

The pose feature $\mathbf{f}(\mathbf{\Theta}^t,\mathbf{v}_c)$ in Eq.~\ref{eq:cano nerf} plays an important role in modeling detailed dynamic human appearance.
Previous methods usually represent the pose feature as low-frequency SMPL-derived attributes (like pose vectors \cite{zheng2022structured} or SMPL positional maps \cite{ma2021power}). However, it is difficult for the network to map these low-frequency driving signals to high-frequency human appearances.
To overcome this challenge, we propose PoseVocab, a novel pose encoding method for realistic human avatar modeling.
Inspired by word embeddings \cite{mikolov2013efficient}, we construct a pose vocabulary, dubbed PoseVocab, which consists of joint-structured key rotations and their corresponding pose embeddings.
PoseVoab encourages the network to discover the optimal pose embeddings to encode the high-frequency dynamic appearance of the character.
Fig.~\ref{fig:overview} demonstrates an overview of the representation of PoseVocab. 
Given the multi-view RGB videos of a character under various poses, we first construct PoseVocab based on the training poses. 
The constructed PoseVocab consists of key rotation samples in $so(3)$ of each joint and the corresponding pose embeddings.
The joint-structured pose embeddings are designed to disentangle the effects of different joints on the pose-dependent dynamic details for better generalization to novel poses.
Given a query pose vector, we first decompose it into rotations of each joint. 
For each query joint rotation, we interpolate joint-structured pose embeddings represented by feature lines through KNN searching and blending.
Finally, the pose-dependent feature of a 3D position is acquired by sampling on the interpolated feature lines for avatar rendering.
Next, we will introduce PoseVocab in detail.

\subsection{Joint-structured Pose Embeddings}
\label{subsec:joint-structured pose embeddings}

The desirable properties of a pose encoding method are effectiveness and generalization. To this end, we propose joint-structured pose embeddings to disentangle the effects of different joints on the dynamic human appearance. 
In other words, we sample key rotations and assign pose embeddings for each joint. 
To be more specific, given the training poses (rotations) $\{\boldsymbol{\theta}_j^t|\boldsymbol{\theta}_j^t\in so(3),1\leq t\leq T\}$ of the $j$-th joint, we first sample $M$ key rotations via farthest point sampling. The distance metric between two rotations is calculated as \cite{tiwari2022pose,huynh2009metrics}:
\begin{equation}
\label{eq:pose dist}
    d(\boldsymbol{\theta}_1,\boldsymbol{\theta}_2)=1-\left|\mathbf{q}(\boldsymbol{\theta}_1)^\top \mathbf{q}(\boldsymbol{\theta}_2)\right|\in [0, 1],
\end{equation}
where $\mathbf{q}(\cdot)$ is a function that maps an axis-angle vector to a unit quaternion, and $\boldsymbol{\theta}_1$ and $\boldsymbol{\theta}_2$ are two axis-angle vectors.
The sampled key rotations $\{\hat{\boldsymbol{\theta}}_j^m|1\leq m\leq M\}$ cover most of the seen poses in the training dataset.
Then we assign a learnable pose embedding for each key rotation.

\begin{figure}[tbp]
    \centering
    \includegraphics[width=\linewidth]{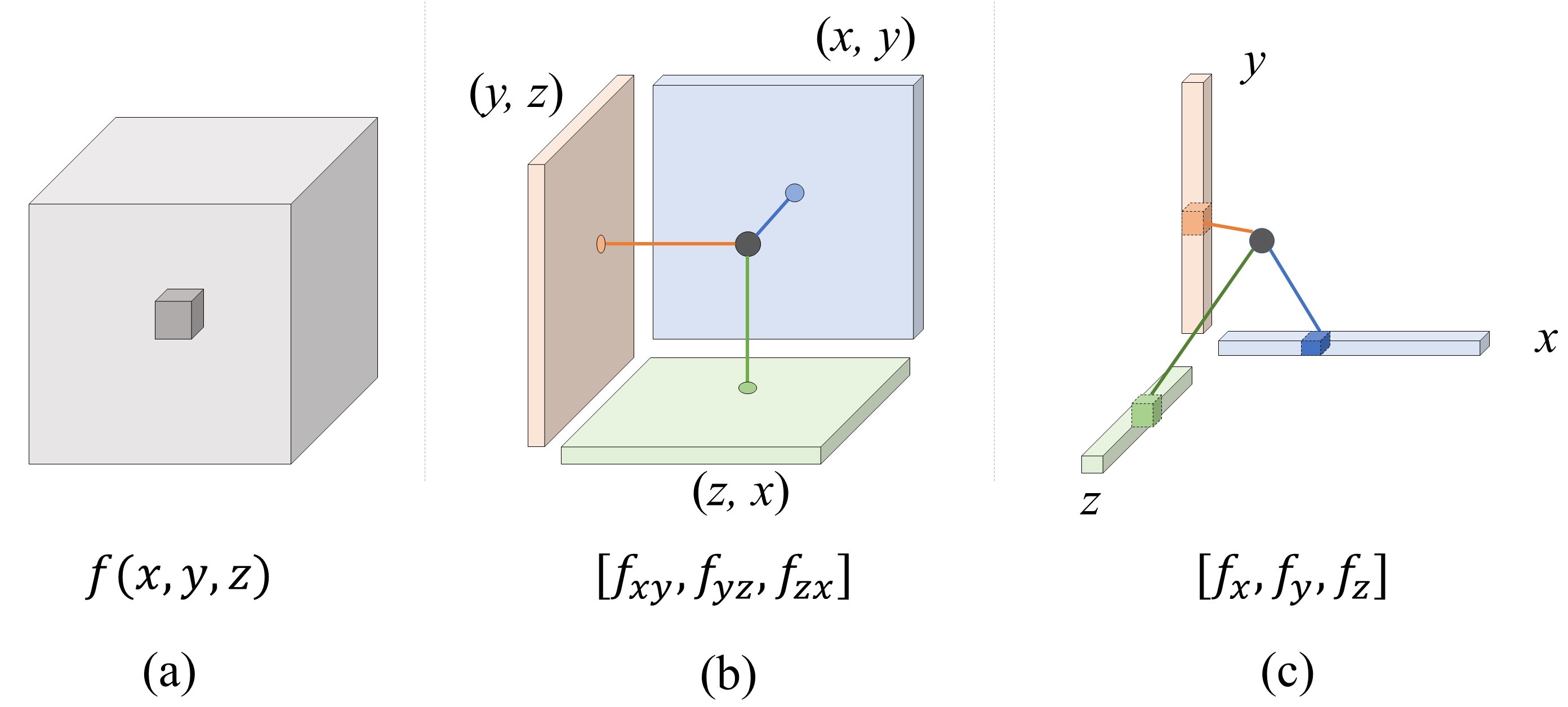}
    \caption{\textbf{Illustrations of volume (a), tri-planes (b) and feature lines (c).} Given a query coordinate $(x,y,z)$, the volume directly returns the feature voxel, tri-planes return the concatenation of three projected features on $(x,y)$, $(y,z)$ and $(z,x)$ planes, and feature lines return the concatenation of three projected features on $x$, $y$ and $z$ axes.}
    \label{fig:feature lines}
\end{figure}

To improve the representation ability of the pose embedding, a possible solution is to represent each pose embedding as volumes \cite{instantngp,fridovich2022plenoxels} or tri-planes \cite{chan2022efficient} (Fig.~\ref{fig:feature lines} (a, b)).
Unfortunately, the memory usage will be unaffordable because the number of pose embeddings is over 5000 in total.
To balance the spatial capacity and memory efficiency, our pose embedding is represented as three feature lines on the $x$, $y$ and $z$ axes.
The structure of feature lines is illustrated in Fig.~\ref{fig:feature lines} (c), and three feature lines are denoted as
\begin{equation}
    \mathbf{F}_{j,x}^m \in \mathbb{R}^{R_x\times D},\,
    \mathbf{F}_{j,y}^m \in \mathbb{R}^{R_y\times D},\,
    \mathbf{F}_{j,z}^m \in \mathbb{R}^{R_z\times D},
    \label{eq:feature lines}
\end{equation}
where $R_x$, $R_y$ and $R_z$ are the resolutions of $x$, $y$ and $z$ feature lines, respectively, and $D$ is the feature dimension. 
Similar to tri-planes \cite{chan2022efficient}, the fetched feature of a 3D point $\mathbf{v}_c=(x_c, y_c, z_c)$ is the concatenation of the projected features on the three lines:
\begin{equation}
\label{eq:sample}
    \begin{split}
    &\mathbf{h}\left(\mathbf{v}_c;F_{j,x}^m,F_{j,y}^m,F_{j,z}^m\right)\\&=\oplus\left(
    \text{Lerp}(x_c;\mathbf{F}_{j,x}^m),
    \text{Lerp}(y_c;\mathbf{F}_{j,y}^m),
    \text{Lerp}(z_c;\mathbf{F}_{j,z}^m)
    \right),
    \end{split}
\end{equation}
where $\oplus$ is the concatenation operation, and $\text{Lerp}(\cdot)$ represents {linear interpolation} on the feature line given a 1D query coordinate.

So far, for each joint, we have constructed $M$ key-value pairs, i.e., key rotations and corresponding pose embeddings, based on the training poses.
These joint-structured pose embeddings assemble the PoseVocab, and serve as discrete samples in the continuous pose feature space for the following query in PoseVocab.

\subsection{Hierarchical Query in PoseVocab}
\label{subsec:hierarchical query}

Based on the constructed PoseVocab that consists of joint-structured key rotations $\{\hat{\boldsymbol{\theta}}_j^m\}$ and pose embeddings $\{\mathbf{F}_{j,x}^m\}$, $\{\mathbf{F}_{j,y}^m\}$, $\{\mathbf{F}_{j,z}^m\}$, we can interpolate these embeddings to acquire the pose feature given an arbitrary pose $\boldsymbol{\Theta}=[\boldsymbol{\theta}_1^\top,...,\boldsymbol{\theta}_J^\top]^\top$ ($J$ is the joint number) and a 3D position.
The query procedure includes three hierarchical levels: joint, key-rotation and spatial levels as shown in Fig.~\ref{fig:overview}. 

\paragraph{Joint Level.} Specifically, given a query pose vector $\boldsymbol{\Theta}=[\boldsymbol{\theta}_1^\top,...,\boldsymbol{\theta}_J^\top]^\top$, we first split it into rotations of each joint $\{\boldsymbol{\theta}_j|1\leq j\leq J\}$.

\paragraph{Key-rotation Level.} For the query rotation of the $j$-th joint, $\boldsymbol{\theta}_j$, we search for the $K$ nearest key rotations $\{\hat{\boldsymbol{\theta}}_j^k\}_{k=1}^K$ using Eq.~\ref{eq:pose dist} as the distance metric, and interpolate corresponding pose embeddings (i.e., feature lines) as a weighted sum:
\begin{equation}
\label{eq:interpolation}
\begin{split}
    \mathbf{F}_{j,x}=\frac{\sum_{k=1}^K w(\boldsymbol{\theta}_j,\hat{\boldsymbol{\theta}}_j^k) \mathbf{F}_{j,x}^k}{\sum_{k=1}^K w(\boldsymbol{\theta}_j,\hat{\boldsymbol{\theta}}_j^k)},
\end{split}
\end{equation}
where $w(\boldsymbol{\theta}_j,\hat{\boldsymbol{\theta}}_j^k)=1-d(\boldsymbol{\theta}_j,\hat{\boldsymbol{\theta}}_j^k)$ is the blending weight. $\mathbf{F}_{j,y}$ and $\mathbf{F}_{j,z}$ are similarly calculated as Eq.~\ref{eq:interpolation}.

\paragraph{Spatial Level.} Given a canonical 3D position $\mathbf{v}_c$, we linearly sample its pose feature $\mathbf{h}(\mathbf{v}_c;\mathbf{F}_{j,x},\mathbf{F}_{j,y},\mathbf{F}_{j,z})$ by Eq.~\ref{eq:sample} on the interpolated pose embeddings, i.e., three feature lines $\mathbf{F}_{j,x}$, $\mathbf{F}_{j,y}$ and $\mathbf{F}_{j,z}$. 
Moreover, following SCANimate \cite{saito2021scanimate}, we also apply a skinning weight awared attention scheme on the feature of each joint to limit the effects of irrelevant joints to reduce spurious correlations.

Finally, we concatenate the features queried by all the joint rotations together as the whole pose feature:
\begin{equation}
    \mathbf{f}(\Theta,\mathbf{v}_c)=\bigoplus_{j=1}^J \left(\omega(\mathbf{v}_c,j)\cdot\mathbf{h}(\mathbf{v}_c;\mathbf{F}_{j,x},\mathbf{F}_{j,y},\mathbf{F}_{j,z})\right),
\end{equation}
where $\omega(\mathbf{v}_c,j)$ is {the predefined influence weight of the $j$-th joint on $\mathbf{v}_c$ \cite{saito2021scanimate}}.
Eventually, the canonical 3D position and its corresponding pose feature are fed into Eq.~\ref{eq:cano nerf} to decode NeRF for avatar rendering.

\subsection{Discussion on PoseVocab Designs}
Our PoseVocab representation is designed for effective and generalized pose encoding based on the following insights:
    
    First of all, it remains difficult for many previous methods \cite{zheng2022structured,li2022tava} to {directly} map SMPL-derived attributes like pose vectors to high-frequency dynamic human appearances {using MLPs because of the low-frequency variation of driving poses and low-frequency bias of MLPs \cite{tancik2020fourier}}.
    In contrast, in our method, the low-frequency pose only plays roles of queries and keys, while the conditional pose feature is the learnable pose embedding.
    We promote the network to discover these pose embeddings to encode high-frequency dynamic human appearances under various poses{, and then input these embeddings containing appearance information into the conditional NeRF MLP.}
    
    Secondly, if we naively construct pairs of global training poses and latent embeddings like \cite{peng2021animatable,wang2022arah}, it still remains difficult to model fine-grained details and generalize to novel poses as shown in Fig.~\ref{fig:eval_hierarchical_query}, because the global pose vector entangles the control of all the joints on the dynamic human appearance.
    So we propose joint-structured pose embeddings to disentangle the effects of different body joints for modeling fine-grained details and generalization ability to novel poses.

    Thirdly, representing each pose embedding as a global latent code yields low-quality avatar appearances without fine-grained patterns of garments as shown in Fig.~\ref{fig:eval_feature_lines}, so we introduce features lines, an effective and compact 3D representation for both representation ability and memory efficiency.
    
    Finally, the hierarchical query in PoseVocab interpolates joint-structured pose embeddings successively in the joint, key-rotation and spatial levels.
    The joint level decomposes the control of different joint motions on the dynamic appearance for better generalization to novel poses.
    The key-rotation level guarantees temporally consistent animation via the smooth KNN blending.
    Last but not least, the spatial level provides the spatial distinction of different positions to encode more fine-grained details.

\subsection{Training}
The learnable variables of our network include the joint-structure pose embeddings $\{\mathbf{F}_{j,x}^m\}$, $\{\mathbf{F}_{j,y}^m\}$, $\{\mathbf{F}_{j,z}^m\}$ and the parameters of the NeRF MLP in Eq.~\ref{eq:cano nerf}.
The training losses include a color loss, a perceptual loss, a mask loss, the Eikonal loss \cite{gropp2020implicit} and a total variation loss on feature lines:
\begin{equation}
    \begin{split}\mathcal{L}&=\lambda_\text{color}\mathcal{L}_\text{color}+\lambda_\text{perceptual}\mathcal{L}_\text{perceptual}\\
    &+\lambda_\text{mask}\mathcal{L}_\text{mask}+\lambda_\text{eikonal}\mathcal{L}_\text{eikonal}+\lambda_\text{TV}\mathcal{L}_\text{TV},
    \end{split}
\end{equation}
where $\lambda$s are loss weights.

\paragraph{Color Loss.} The color loss is an L1 loss between the volume-rendered {\cite{yariv2021volume}} color image and the ground truth:
\begin{equation}
    \mathcal{L}_\text{color}=\sum_{\mathbf{r}\in\mathcal{R}}\left\| \mathbf{C}(\mathbf{r})-\mathbf{C}^*(\mathbf{r})\right\|_1,
\end{equation}
where $\mathcal{R}$ is the set of sampled rays from the rendered view, and $\mathbf{C}(\mathbf{r})$ and $\mathbf{C}^*(\mathbf{r})$ are the rendered and true pixel colors, respectively.

\paragraph{Perceptual Loss.} The perceptual loss \cite{zhang2018unreasonable} is conducted on a local patch of rendered and ground-truth images, and penalizes them to be close in the feature map level. We choose VGG as the backbone to compute the learned perceptual image patch similarity (LPIPS):
\begin{equation}
    \mathcal{L}_\text{perceptual}=\sum_{\mathbf{p}\in\mathcal{P}}\left\|\text{VGG}(\mathbf{C}(\mathbf{p}))-\text{VGG}(\mathbf{C}^*(\mathbf{p}))\right\|_2^2,
\end{equation}
where $\mathcal{P}$ is the set of sampled patches from the rendered view, and $\mathbf{C}(\mathbf{p})$ and $\mathbf{C}^*(\mathbf{p})$ are the rendered and true patches, respectively. The perceptual loss has been widely used in the NeRF training \cite{weng2022humannerf,gao2022reconstructing} to improve the reconstructed details.

\paragraph{Mask Loss.} The mask loss constrains the rendered mask to be consistent with the ground truth:
\begin{equation}
    \mathcal{L}_\text{mask}=\sum_{\mathbf{r}\in\mathcal{R}}\left\|M(\mathbf{r})-M^*(\mathbf{r})\right\|_1,
\end{equation}
where $M(\mathbf{r})$ and $M^*(\mathbf{r})$ are volume-rendered and ground-truth mask values, respectively. 
The mask loss enforces the modeled human geometry to be consistent with the 2D body mask.

\paragraph{Eikonal Loss.} The Eikonal loss \cite{gropp2020implicit} is an implicit geometric regularization that enforces the norm of the gradient of the SDF field equal to 1:
\begin{equation}
    \mathcal{L}_\text{eikonal}=\mathbb{E}\left(\|\nabla_\mathbf{v}s(\mathbf{v},\mathbf{f}(\Theta,\mathbf{v}))\|_2^2 - 1\right),
\end{equation}
where $s(\cdot)$ is the MLP-based function that maps a 3D position $\mathbf{v}$ and its conditional pose feature $\mathbf{f}(\Theta,\mathbf{v})$ to the SDF value.

\paragraph{Total Variation Loss.} The total variation (TV) loss regularizes the continuity of the feature lines along the spatial dimension:
\begin{equation}
\begin{split}
    \mathcal{L}_\text{TV}&=\sum_{j,m}\sum_i\left(
    \left\|\mathbf{F}_{j,x}^m(i+1)-\mathbf{F}_{j,x}^m(i)\right\|_2^2+
    \left\|\mathbf{F}_{j,y}^m(i+1)-\mathbf{F}_{j,y}^m(i)\right\|_2^2\right.\\&\left.+
    \left\|\mathbf{F}_{j,z}^m(i+1)-\mathbf{F}_{j,z}^m(i)\right\|_2^2\right),
\end{split}
\end{equation}
where $i$ is the index on the spatial dimension of feature lines.

%% file: secs/4_results.tex
\section{Experiments}

\paragraph{Dataset.}
We use 5 multi-view sequences for the experiments: 3 sequences with 24 views from THuman4.0 dataset \cite{zheng2022structured}, 1 sequence with 11 views from DeepCap dataset \cite{habermann2020deepcap} and 1 sequence with 23 views from ZJU-MoCap dataset \cite{peng2021neural}.
All the sequences provide SMPL \cite{loper2015smpl} or SMPL-X \cite{SMPL-X:2019} registrations of the character.
{We split each sequence into two continuous chunks for training and testing, and the training chunk accounts for $50\% \sim 80\%$. Novel poses are from the testing chunk or another sequence.}

\paragraph{Metric.}
We utilize Peak Signal-to-Noise Ratio (PSNR), Structure Similarity Index Measure (SSIM) \cite{wang2004image}, Learned Perceptual Image Patch
Similarity (LPIPS) \cite{zhang2018unreasonable} and Frechet Inception Distance (FID) \cite{heusel2017gans} as metrics for quantitative comparisons and evaluations.

\subsection{Results}
We train the avatar network for each subject individually and demonstrate results animated by novel poses in Fig.~\ref{fig:teaser} and Fig.~\ref{fig:results}.
Our results show realistic dynamic details varying with the driving pose.
Please refer to the supplemental video for more visualization of the animatable avatars.

\input{tabs/tab_comparison.tex}
\input{tabs/tab_comparison_na.tex}

\subsection{Comparison}
{We mainly compare our method against 5 state-of-the-art approaches including Structured Local NeRF (SLRF) \cite{zheng2022structured}, TAVA \cite{li2022tava}, ARAH \cite{wang2022arah}, Animatable NeRF (Ani-NeRF) \cite{peng2021animatable} and Neural Actor (NA) \cite{liu2021neural} on the quality of animated avatars.}

\paragraph{SLRF, TAVA, ARAH and Ani-NeRF}
Fig.~\ref{fig:cmp_slrf} shows qualitative comparisons against SLRF, TAVA and ARAH on DeepCap \cite{habermann2020deepcap} and THuman4.0 \cite{zheng2022structured} datasets.
We also compare our method against Ani-NeRF on ZJU-MoCap dataset \cite{peng2021neural} in Fig.~\ref{fig:cmp_aninerf}.
The animated results by SLRF, TAVA, ARAH and Ani-NeRF are blurry especially in the red circles, whereas our method not only reconstructs more details in terms of garment wrinkles under the training poses, but also generates more fine-grained and realistic dynamic appearance given novel poses.
Although SLRF represents the avatar as a set of local radiance fields to improve the network capacity, it still suffers from the bottleneck of its pose encoding, i.e., the low-frequency pose vector.
Moreover, neither the learned pose-dependent shading in TAVA nor per-frame latent codes in Ani-NeRF and ARAH are able to model detailed dynamic human appearance.
Tab.~\ref{tab:cmp} reports the numerical comparison on the animation accuracy.
Overall, our method outperforms these four approaches both qualitatively and quantitatively benefiting from the proposed pose encoding method, PoseVocab, which has a powerful ability to encode high-frequency dynamic human appearance.

{\paragraph{Neural Actor (NA)}
We compare our method with NA both qualitatively and quantitatively on ``S2'' sequence of DeepCap dataset in Fig.~\ref{fig:cmp_na} and Tab.~\ref{tab:cmp_na}, respectively.
We follow the same training/testing splits and metric computation as NA, and the visualized and quantitative results are borrowed from \cite{liu2021neural}.
On the whole, NA outperforms the other four SOTA methods (SLRF, TAVA, etc), and the possible reason is that it transforms the 3D human surface into 2D SMPL UV to utilize a powerful 2D image-to-image translation network with adversarial training to predict texture maps from normal maps.
Without deep 2D convolutions, our method can produce comparable or even better results than NA as shown in Fig.~\ref{fig:cmp_na}.
Moreover, the SMPL UV parameterization in NA constrains the character to wear tight clothes, while PoseVocab is a basic pose encoding method that can be combined with other avatar representations for modeling loose clothes as discussed in Sec.~\ref{sec:discussion}.
}

\subsection{Evaluation}
We conduct evaluations on the core designs of our method to demonstrate the improvement brought by our contributions.

\begin{figure}[t]
    \centering
    \includegraphics[width=\linewidth]{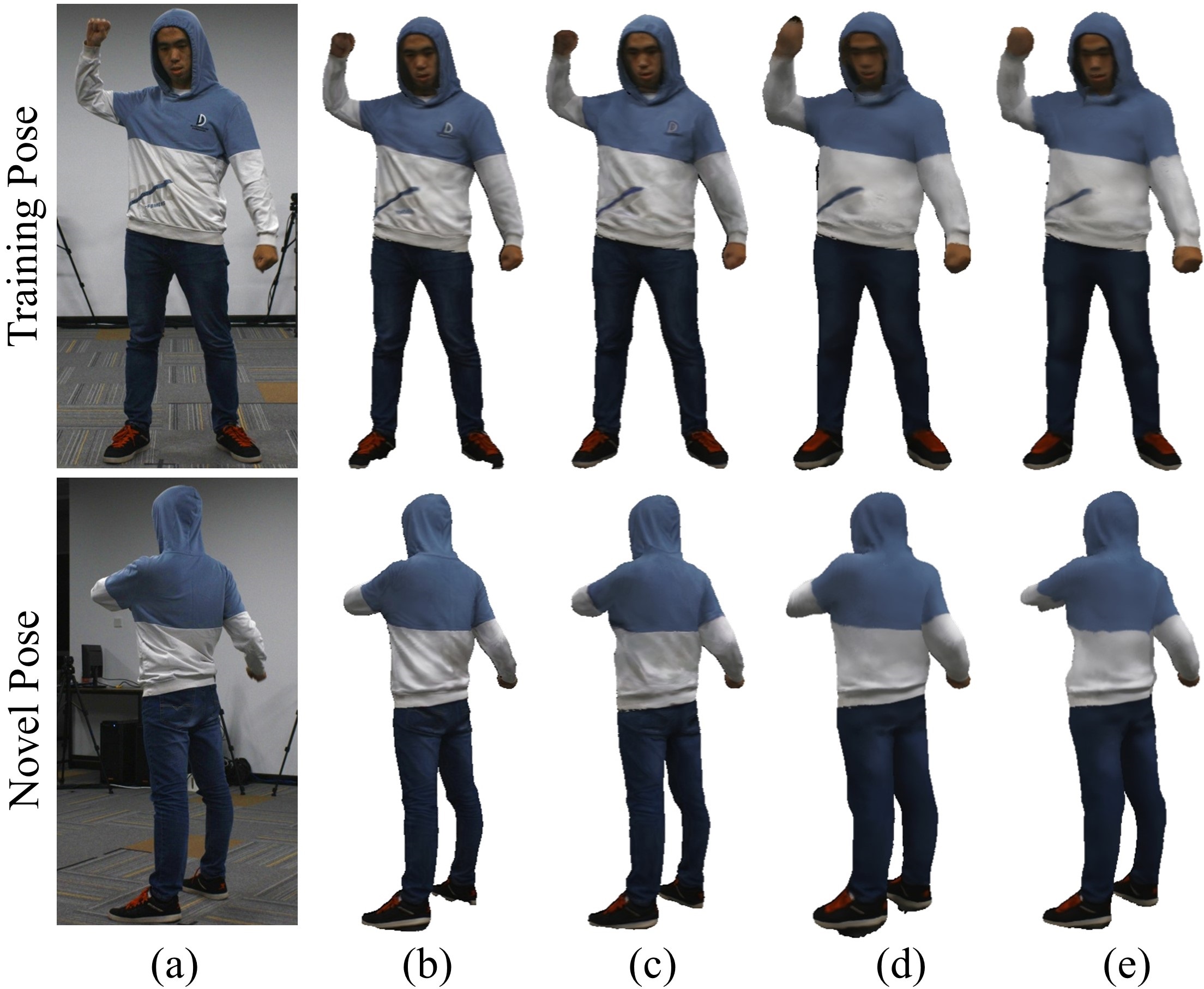}
    \caption{{\textbf{Qualitative evaluation on pose encoding methods.} 
    We show ground-truth images (a), and synthesized avatars by PoseVocab (b), pose vectors (c) and SMPL positional (d) and normal (e) maps, respectively.
    }}
    \label{fig:eval_pose_encoding}
\end{figure}

\paragraph{Pose Encoding Methods.}
{To prove the effectiveness of our pose encoding method, PoseVocab, we compare it against 3 baseline methods, i.e., \textit{pose vectors} \cite{zheng2022structured}, \textit{SMPL positional maps} \cite{ma2021power} and \textit{SMPL normal maps} \cite{yoon2022learning}.}
Specifically, the three baseline methods respectively take the pose vector, convolutional SMPL positional or normal feature as the pose condition to decode the dynamic details of the character.
Note that since all the points in the canonical 3D space require corresponding pose features, we sample them by orthographic projection on rendered maps as in \cite{li2022avatarcap,lin2022learning}.
{Fig.~\ref{fig:eval_pose_encoding} shows that the proposed PoseVocab not only encodes more fine-grained details under the training poses, but also generates more realistic appearances in novel poses than the other three pose encoding methods.}
Tab.~\ref{tab:eval_pose_encoding} shows that our method can produce more accurate results on both novel view and novel pose synthesis.

\input{tabs/tab_pose_encoding.tex}

\begin{figure}[t]
    \centering
    \includegraphics[width=\linewidth]{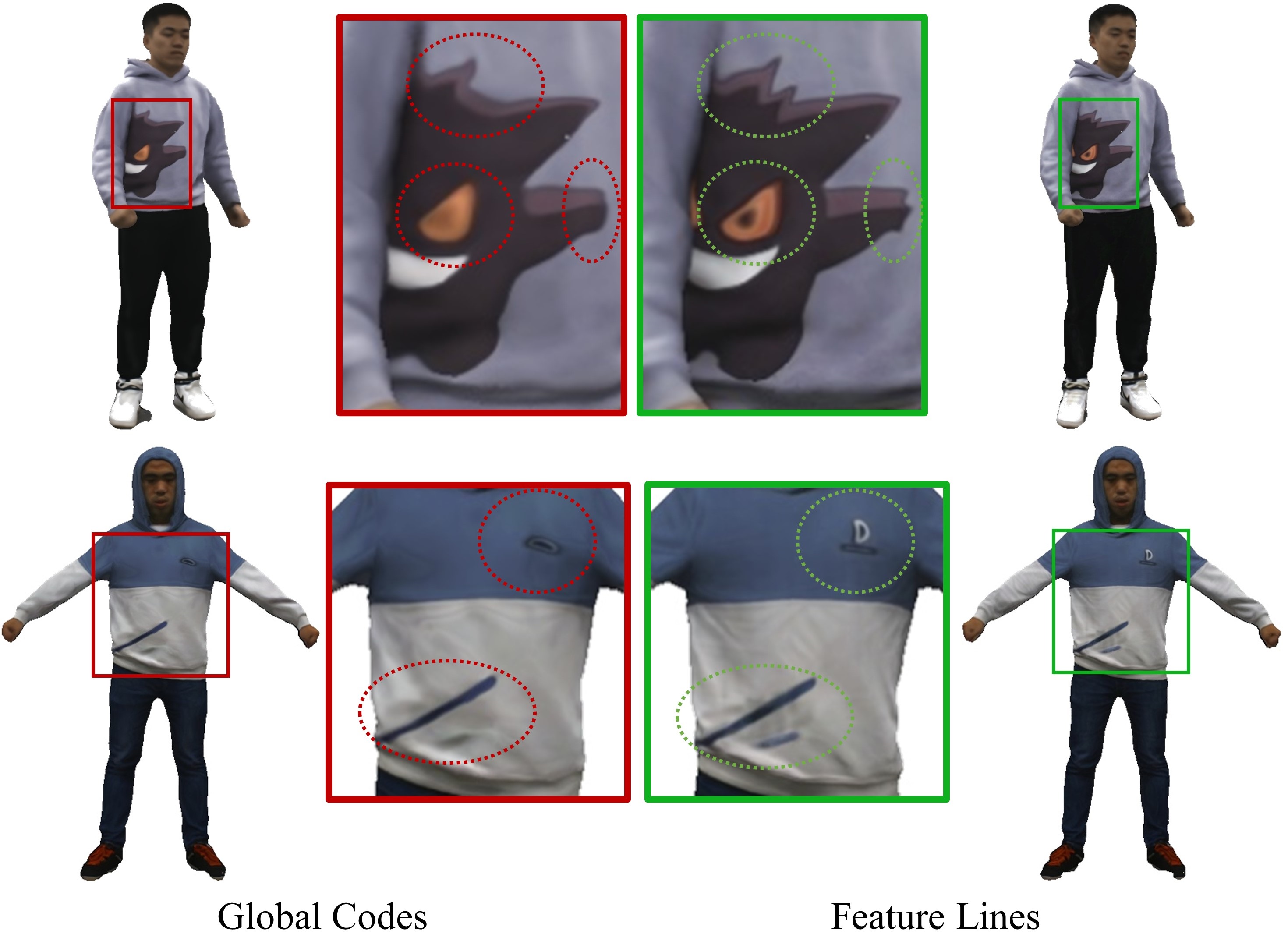}
    \caption{\textbf{Qualitative ablation study on Feature lines.}
    Rendering results of joint-structured pose embeddings represented by global codes and feature lines, respectively.}
    \label{fig:eval_feature_lines}
\end{figure}

\begin{figure}[t]
    \centering
    \includegraphics[width=\linewidth]{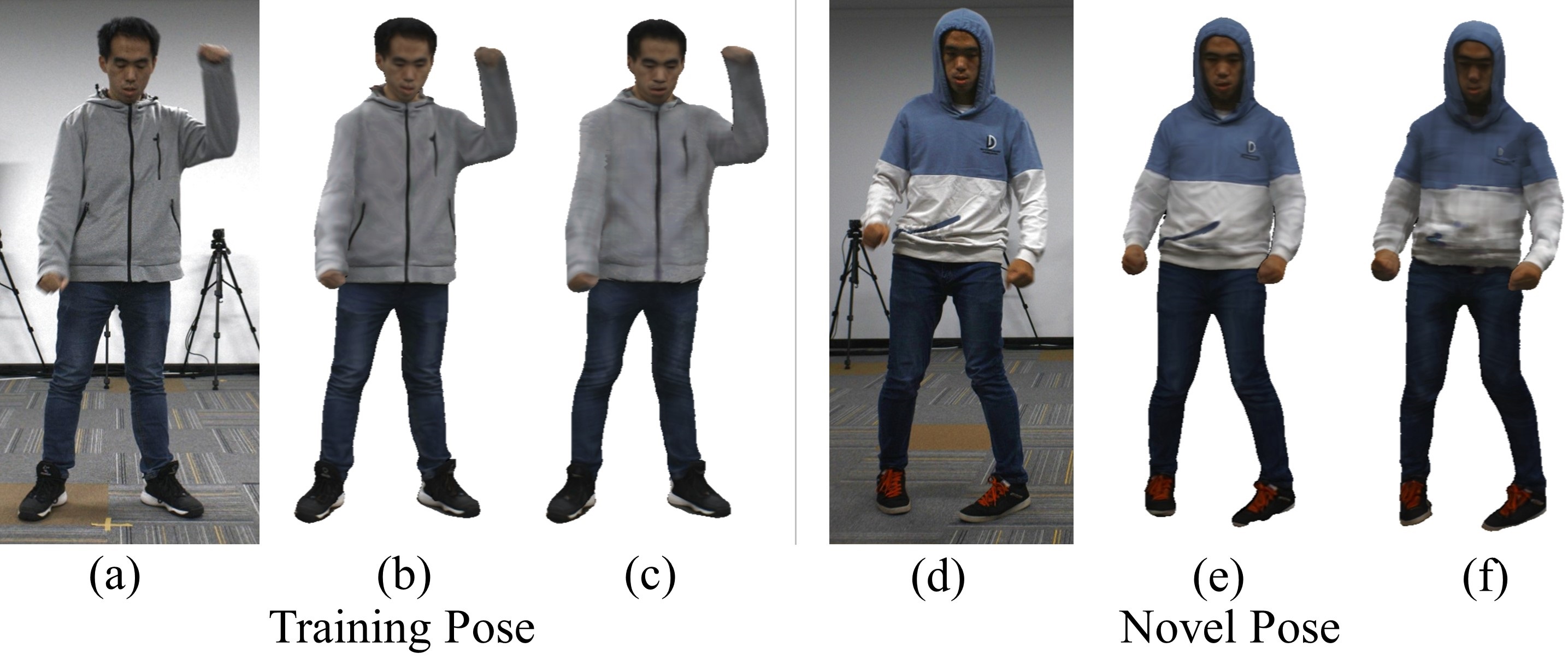}
    \caption{{\textbf{Qualitative ablation study on the joint level of the hierarchical query.}
    (a)(d) Ground truth, (b)(e) results with the joint level, (c)(f) results without the joint level.}}
    \label{fig:eval_hierarchical_query}
\end{figure}

\paragraph{Ablation Study on Feature Lines.}
In our method, we propose to represent each pose embedding as three feature lines on $x$, $y$ and $z$ axes to improve the spatial capacity.
We evaluate the capacity of feature lines by replacing them with global latent codes, i.e., all the 3D points in the canonical space share the same pose embedding.
Fig.~\ref{fig:eval_feature_lines} shows the rendering results with global latent codes and feature lines, respectively.
It demonstrates that the proposed feature lines have a more powerful ability to preserve fine-grained details of human appearances, especially patterns and logos on the clothes.
On the other hand, the model sizes of networks with feature lines and global codes are 194.0 and 6.9 MB, respectively, and are both affordable for commercial GPUs.
{However, representing each pose embedding as tri-planes \cite{chan2022efficient} or volumes by hash tables \cite{instantngp} is unaffordable since the embedding number is very large (over 5000).}
Overall, the proposed feature lines are both effective and memory-efficient.

\paragraph{Ablation Study on Joint Level in Hierarchical Query.}
The hierarchical query in PoseVocab includes joint, key-rotation and spatial levels.
We attempt to eliminate the joint level and redivide the query procedure into key-pose and spatial levels, i.e., we sample key global pose vectors (key poses) rather than key rotations of each joint, and then assign a pose embedding for each key pose.
In this setting, given a query pose, we directly search for $K$ nearest key poses and interpolate corresponding embeddings to acquire the pose feature.
We show animated results with and without the joint level in Fig.~\ref{fig:eval_hierarchical_query}, and it demonstrates that the joint level not only reconstructs more fine-grained details under the training pose, but also improves the generalization ability to novel poses by disentangling the effects of different joints on the dynamic human appearance.

{\paragraph{Pose Generalization.}
We evaluate the pose generalization of PoseVocab using driving poses with different pose similarities from the testing chunk and AIST++ dataset \cite{li2021ai}.
We also evaluate the results of a baseline method that replaces PoseVocab with continuous MLPs to model the mapping from driving poses to latent embeddings.
Fig.~\ref{fig:eval_pose_generalization} shows that PoseVocab outperforms the baseline method on pose generalization and avatar quality.
The reason for our superiority may be that PoseVocab implicitly projects the novel driving pose to the pose space spanned by seen poses instead of extrapolating by a learned MLP-based function.
}

%% file: tabs/tab_comparison.tex
\begin{table}[t]
\caption{{\textbf{Quantitative comparison against SLRF, Ani-NeRF, TAVA and ARAH on ``subject00'' sequence of THuman4.0 dataset.} Metrics are computed on both training and testing poses.}}
\begin{tabular}{llcccc}
\hline
Pose                & Method  & PSNR $\uparrow$  & SSIM $\uparrow$ & LPIPS $\downarrow$ & FID $\downarrow$ \\ \hline
\multirow{4}{*}{Training} & Ours           & \textbf{34.226} & \textbf{0.986} & \textbf{0.014}     & \textbf{23.957}  \\
                          & SLRF    &     25.270      &      0.971     &    0.024        &      44.492      \\
                          & Ani-NeRF &    23.188       &     0.966      &          0.033     &    85.449       \\
                          & TAVA     &    23.934       &     0.967      &          0.029     &     75.464      \\ 
                          & ARAH & 22.017 & 0.963 & 0.033 & 74.308 \\
                          \hline
\multirow{4}{*}{Novel}    & Ours           & \textbf{30.972} & \textbf{0.977} & \textbf{0.017}     & \textbf{37.239}  \\
                          & SLRF    &      26.152     &     0.969      &        0.024       &     110.651      \\
                          & Ani-NeRF &    22.532      &     0.964    &     0.034        &         102.233 \\
                          & TAVA     &      26.607     &      0.968     &       0.032      &     99.947     \\ 
                          & ARAH & 21.769 & 0.958 & 0.037 & 77.840 \\
                          \hline
\end{tabular}
\label{tab:cmp}
\end{table}

%% file: tabs/tab_comparison_na.tex
\begin{table}[t]
\caption{{
\textbf{Quantitative comparison against Neural Actor on ``S2'' sequence of DeepCap dataset.}
Metrics are computed on the testing sequence in the same cropped manner as \cite{liu2021neural}.
}}
\begin{tabular}{llll}
\hline
Method       & \multicolumn{1}{c}{PSNR $\uparrow$} & \multicolumn{1}{c}{LPIPS $\downarrow$} & \multicolumn{1}{c}{FID $\downarrow$} \\ \hline
Our          & \textbf{25.836}                     & \textbf{0.061}                         & \textbf{15.228}                      \\
Neural Actor & 23.531                              & 0.066                                  & 19.714                               \\ \hline
\end{tabular}
\label{tab:cmp_na}
\end{table}

%% file: tabs/tab_pose_encoding.tex
\begin{table}[t]
\caption{\textbf{Quantitative evaluation on pose encoding methods.} Numerical results on both training and novel poses by PoseVocab (ours), pose vectors, SMPL positional and normal maps, respectively.}
\begin{tabular}{llcccc}
\hline
Pose                & Method  & PSNR $\uparrow$  & SSIM $\uparrow$ & LPIPS $\downarrow$ & FID $\downarrow$ \\ \hline
\multirow{4}{*}{Training} & Ours           & \textbf{27.958} & \textbf{0.980} & \textbf{0.016}     & \textbf{45.486}  \\
                          & Pose Vector    & 25.848          & 0.974          & 0.024              & 48.965           \\
                          & Pos. Map & 24.931          & 0.971          & 0.027              & 71.121           \\
                          & Norm. Map     & 25.330          & 0.971          & 0.027              & 67.417           \\ \hline
\multirow{4}{*}{Novel}    & Ours           & \textbf{27.464} & \textbf{0.979} & \textbf{0.014}     & \textbf{64.396}  \\
                          & Pose Vector    & 25.384          & 0.977          & 0.020              & 79.256           \\
                          & Pos. Map & 23.201          & 0.973          & 0.025              & 85.226           \\
                          & Norm. Map     & 25.323          & 0.977          & 0.021             & 80.389           \\ \hline
\end{tabular}
\label{tab:eval_pose_encoding}
\end{table}

%% file: secs/5_discussion.tex
\section{Discussion}
\label{sec:discussion}
\paragraph{Conclusion.}
In this paper, we present PoseVocab, a novel pose encoding method for human avatar modeling.
We propose joint-structured pose embeddings to encode the dynamic human appearance under various body poses.
Compared with previous methods that directly map the low-frequency SMPL-derived attributes like pose vectors \cite{zheng2022structured,li2022tava} to the high-frequency dynamic human appearances, our approach promotes the network to discover the optimal pose embeddings to encode the high-fidelity varying details of the avatar.
Furthermore, we introduce feature lines to improve the representation ability of the pose embedding while maintaining memory efficiency.
Last but not least, a hierarchical query strategy in PoseVocab is designed for disentangling the control of different joints on the dynamic human appearance and for generalized and temporally consistent avatar animation.
Overall, our approach outperforms other state-of-the-art methods both qualitatively and quantitatively, and we believe that the proposed new pose encoding method will make progress towards realistic animatable human avatar modeling.

\paragraph{Limitation.}
So far, our method cannot handle the character wearing loose clothes like long dresses, because our avatar representation relies on the inverse skinning by SMPL skeletons.
But the proposed PoseVocab is a general pose encoding method, and we believe that it can be applied to other avatar representations, e.g, DDC \cite{habermann2021real} and SLRF \cite{zheng2022structured}, to model the character wearing loose clothes.

{\paragraph{Social Impact.} Our method can automatically create animatable human avatars, and may be combined with Deep Fakes to generate fake videos. This should be addressed carefully before deploying the technology.}

\begin{acks}
This paper is supported by National Key R\&D Program of China (2022YFF0902200), the NSFC project No.62125107.
\end{acks}

%% file: secs/7_supp.tex
\appendix
In this supplementary material, we will introduce the implementation, training, and animation details.

\section{Implementation Details}
\begin{figure}[t]
    \centering
    \includegraphics[width=\linewidth]{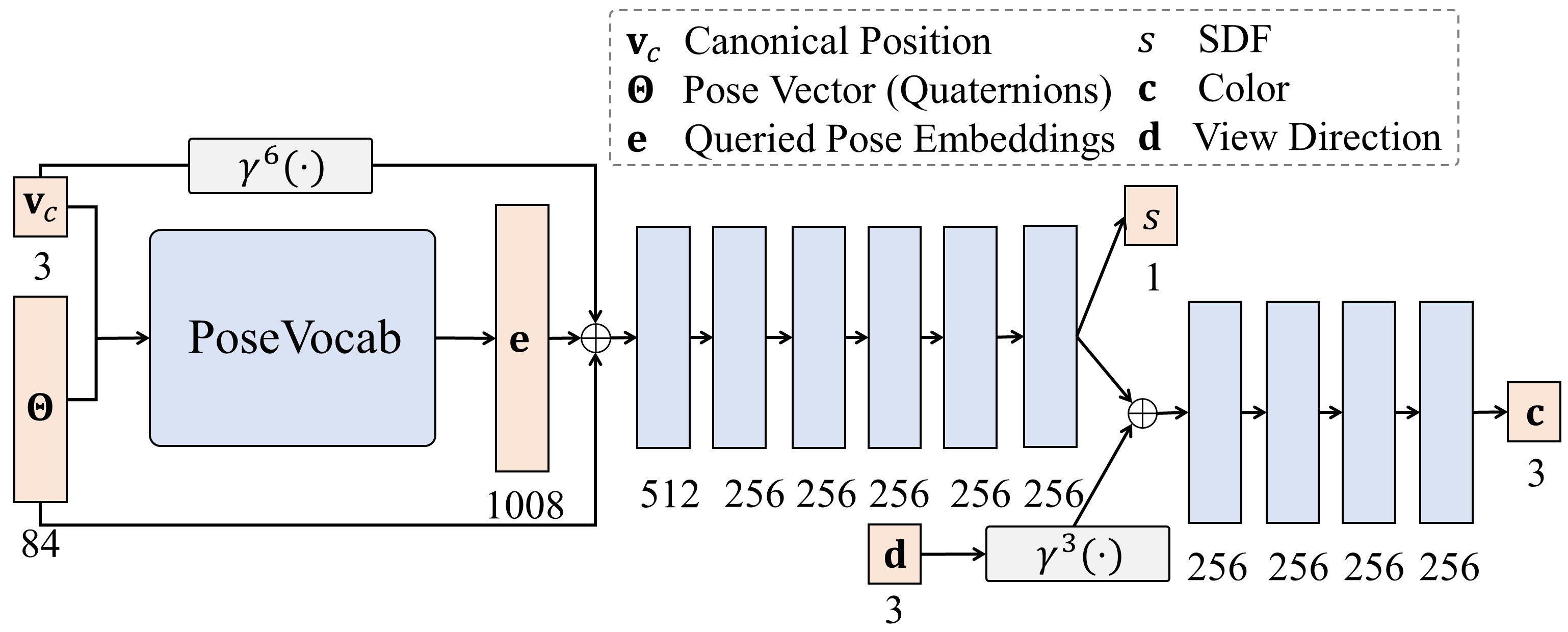}
    \caption{\textbf{Network architecture.}}
    \label{fig:net_arch}
\end{figure}

We implement the network by PyTorch \cite{paszke2019pytorch}. To improve the spatial continuity, we construct PoseVocab on multiple spatial scales like Instant-NGP \cite{instantngp}, and concatenate the embeddings queried on each scale together. The hyper-parameters of PoseVocab are listed in Tab.~\ref{tab:hyper-parameters}. The NeRF \cite{mildenhall2020nerf} module is instantiated as an MLP. The whole network architecture is illustrated in Fig.~\ref{fig:net_arch}.
The frequency of positional encoding \cite{tancik2020fourier} for the position and view direction is 6 and 3, respectively.
The pose vector is represented as quaternions of 21 SMPL joints except for the root and hand joints, since the global rotation should not affect the avatar appearance in most situations and we do not focus on modeling the hands.
The pose vector is also concatenated with queried embeddings and fed into the MLP.

\section{Training Details}
\label{sec:training_details}
We train the network using the Adam \cite{adam} optimizer with a batch size of 1 for 50 epochs.
The initial learning rate is $5\times 10^{-4}$ and decays by multiplying $0.8$ every 100K iterations.
The training procedure takes about $1.5\sim 2$ days on one RTX 3090 card.
The training procedure contains three stages.
In the first stage, we set $\lambda_\text{color}=1$, $\lambda_\text{perceptual}=0$, $\lambda_\text{mask}=1$, $\lambda_\text{eikonal}=0.1$ and $\lambda_\text{TV}=10$ during the first 5 epochs. 
We randomly sample 1024 rays on the training views and sample 64 points on each ray within the SMPL bounding box.
Under the supervision of the mask and color losses, a plausible geometry for each training frame can be learned after 5 epochs.
Then we extract 3D meshes using Marching Cubes \cite{lorensen1987marching} for all the training frames by querying the SDF value of each voxel in a coarse 3D volume that contains the posed character, and then render depth maps to all the training views.
In the following training procedure, we only sample 32 points within 5 cm near the surface based on the rendered depth map.
The depth-guided sampling strategy encourages the network to focus on modeling the dynamic appearance of valid regions.
In the second stage, we disable the Eikonal loss for faster training from the 5th to the 30th epoch. 
In the third stage, we sample patches with a resolution of $64\times 64$ on the training view, and enable the perceptual loss with $\lambda_\text{perceptual}$ set to 0.1 until convergence at the 50th epoch.

\section{Animation Details}
\label{sec:animation_details}
Given a testing pose from the testing chunk or another dataset, we first obtain the SMPL model and allocate a sparse volume with a resolution of $128\times 128\times 128$ that contains the posed SMPL. 
Then we predict SDF values of voxels near the SMPL surface to extract the geometric avatar using Marching Cubes \cite{lorensen1987marching}.
With the geometric model on hand, we can render it to the camera view to obtain a depth map.
In the next volume rendering, we only need to evaluate the colors of pixels inside the body mask on the depth map. 
Besides, we can sample points near the geometric surface based on the depth map in volume rendering for clearer texture.
But the depth may be inaccurate on boundaries of self-occluded regions, producing background colors (e.g., black or white).
So we reevaluate these pixels by sampling points within the balls generated by SMPL vertices like \cite{liu2021neural}.
Such a rendering strategy improves the inference speed so that it can take about 3 secs to render the neural avatar at a resolution of $512\times 512$.
In sequential animation, a sliding window of length 5 is introduced to jointly consider embeddings of adjacent frames for more stable temporal consistency.

\section{Geometric Results}
\begin{figure}[t]
    \centering
    \includegraphics[width=0.96\linewidth]{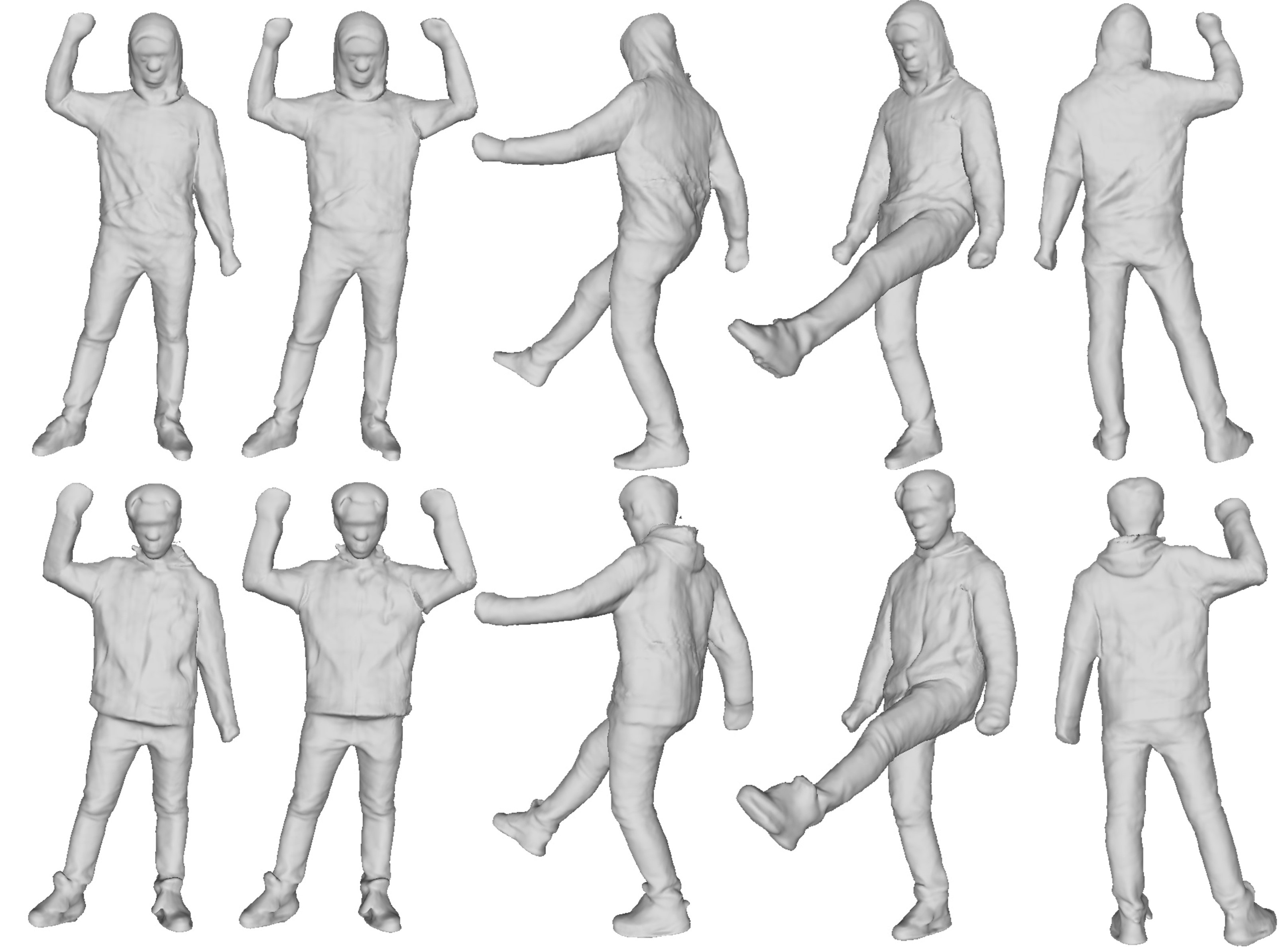}
    \caption{\textbf{Example geometric avatars.}}
    \label{fig:geo_results}
\end{figure}

Benefiting from SDF-based geometric representation \cite{yariv2021volume} and the effectiveness of PoseVocab, our method can also produce detailed geometric avatars as shown in Fig.~\ref{fig:geo_results}.
As mentioned in Sec.~\ref{sec:training_details} and Sec.~\ref{sec:animation_details}, the geometry serves as a prior to sample points near the surface in volume rendering.

\input{tabs/tab_hyperparameters.tex}

%% file: tabs/tab_hyperparameters.tex
\begin{table}[t]
\caption{\textbf{Hyper-parameters of PoseVocab module.} We implement the PoseVocab module on 4 multiple spatial scales, and report its hyper-parameters on each scale, respectively.}
\begin{tabular}{lc}
\hline
Parameter & Value  \\ \hline
$M$ (Number of Key Rotations) & (256, 256, 256, 256) \\
$R_x,R_y,R_z$ in Eq.~3
(Spatial Resolution) & (256, 128, 32, 8) \\
$D$ in Eq.~3
(Feature Channels) & (4, 4, 4, 4)\\
$K$ in Eq.~5
(KNN Number) & (8, 8, 8, 8) \\
\hline
\end{tabular}
\label{tab:hyper-parameters}
\end{table}

%% file: secs/6_figure_only.tex
\clearpage
\begin{figure*}[t]
    \centering
    \includegraphics[width=\linewidth]{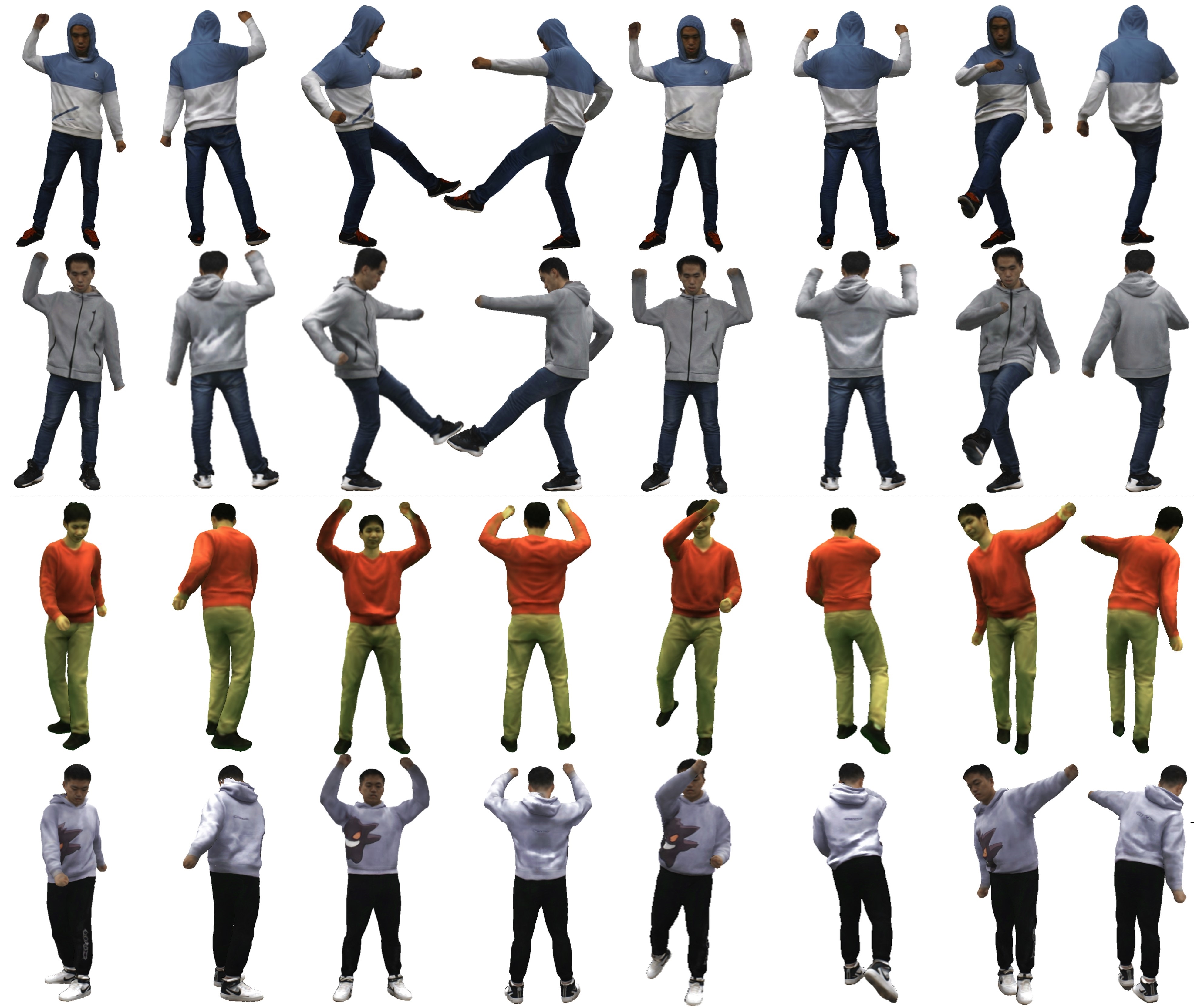}
    \caption{Animated avatars with high-fidelity pose-dependent dynamic details by our method.}
    \label{fig:results}
\end{figure*}

\begin{figure*}[t]
\begin{minipage}{0.48\linewidth}
    \centering
    \includegraphics[width=\linewidth]{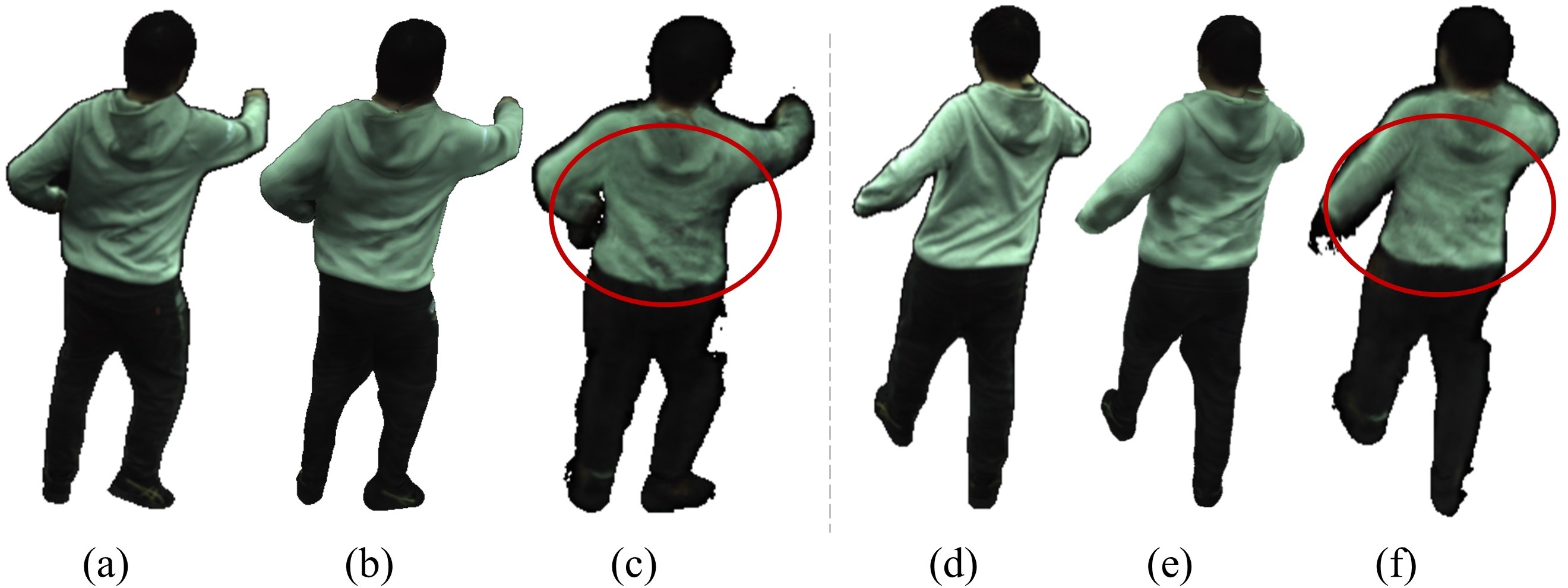}
    \caption{{\textbf{Qualitative comparison against Ani-NeRF \cite{peng2021animatable} under both training (left) and testing (right) poses.} (a)(d) Ground truth, (b)(e) our results, (c)(f) results of Ani-NeRF.}}
    \label{fig:cmp_aninerf}
\end{minipage}
\hfill
\begin{minipage}{0.48\linewidth}
    \centering
    \includegraphics[width=\linewidth]{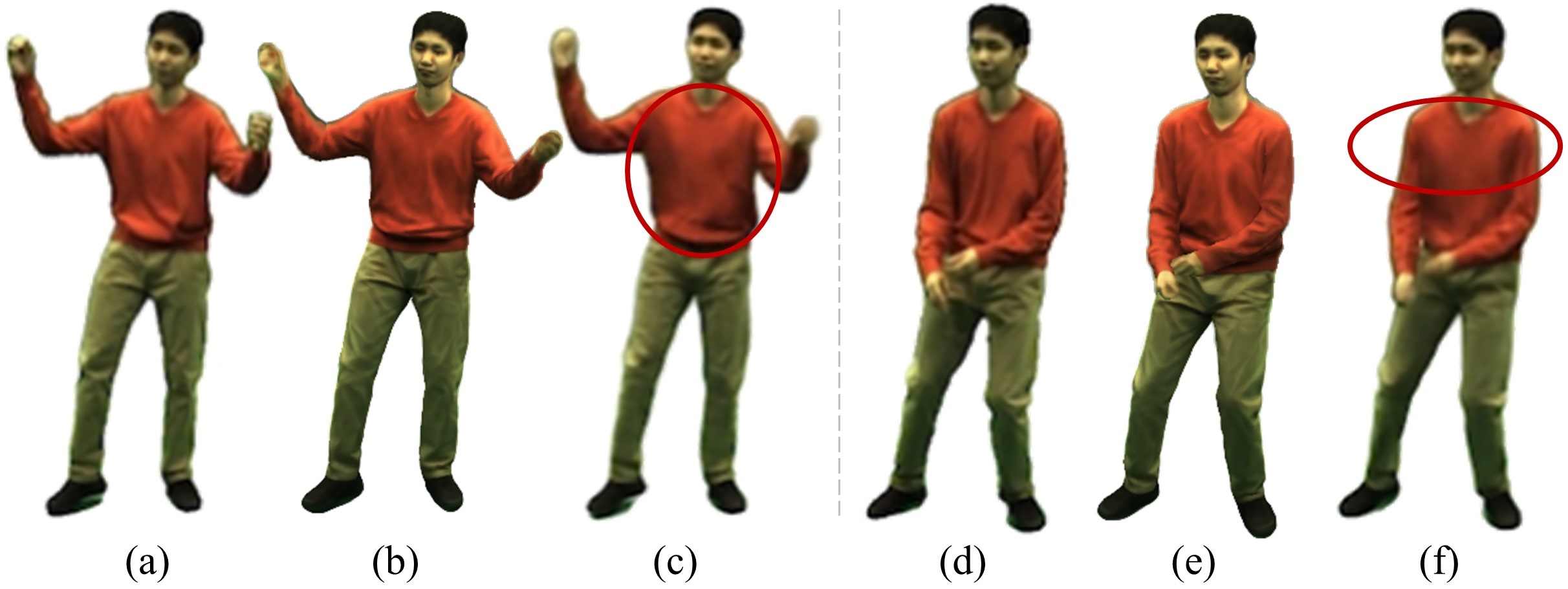}
    \caption{{\textbf{Qualitative comparison against Neural Actor \cite{liu2021neural} on novel pose synthesis.}
    (a)(d) Ground truth, (b)(e) Our results, (c)(f) results of Neural Actor.}}
    \label{fig:cmp_na}
\end{minipage}
\end{figure*}

% \begin{figure}[t]
%     \centering
%     \includegraphics[width=\linewidth]{figs/cmp_aninerf.jpg}
%     \caption{{\textbf{Qualitative comparison against Ani-NeRF \cite{peng2021animatable} under both training (left) and testing (right) poses.} (a)(d) Ground truth, (b)(e) our results, (c)(f) results of Ani-NeRF.}}
%     \label{fig:cmp_aninerf}
% \end{figure}

% \begin{figure}[t]
%     \centering
%     \includegraphics[width=\linewidth]{figs/cmp_na.jpg}
%     \caption{{\textbf{Qualitative comparison against Neural Actor \cite{liu2021neural} on novel pose synthesis.}
%     (a)(d) Ground truth, (b)(e) Our results, (c)(f) results of Neural Actor.}}
%     \label{fig:cmp_na}
% \end{figure}

\begin{figure*}[t]
\begin{minipage}{0.48\linewidth}
    \centering
    \includegraphics[width=\linewidth]{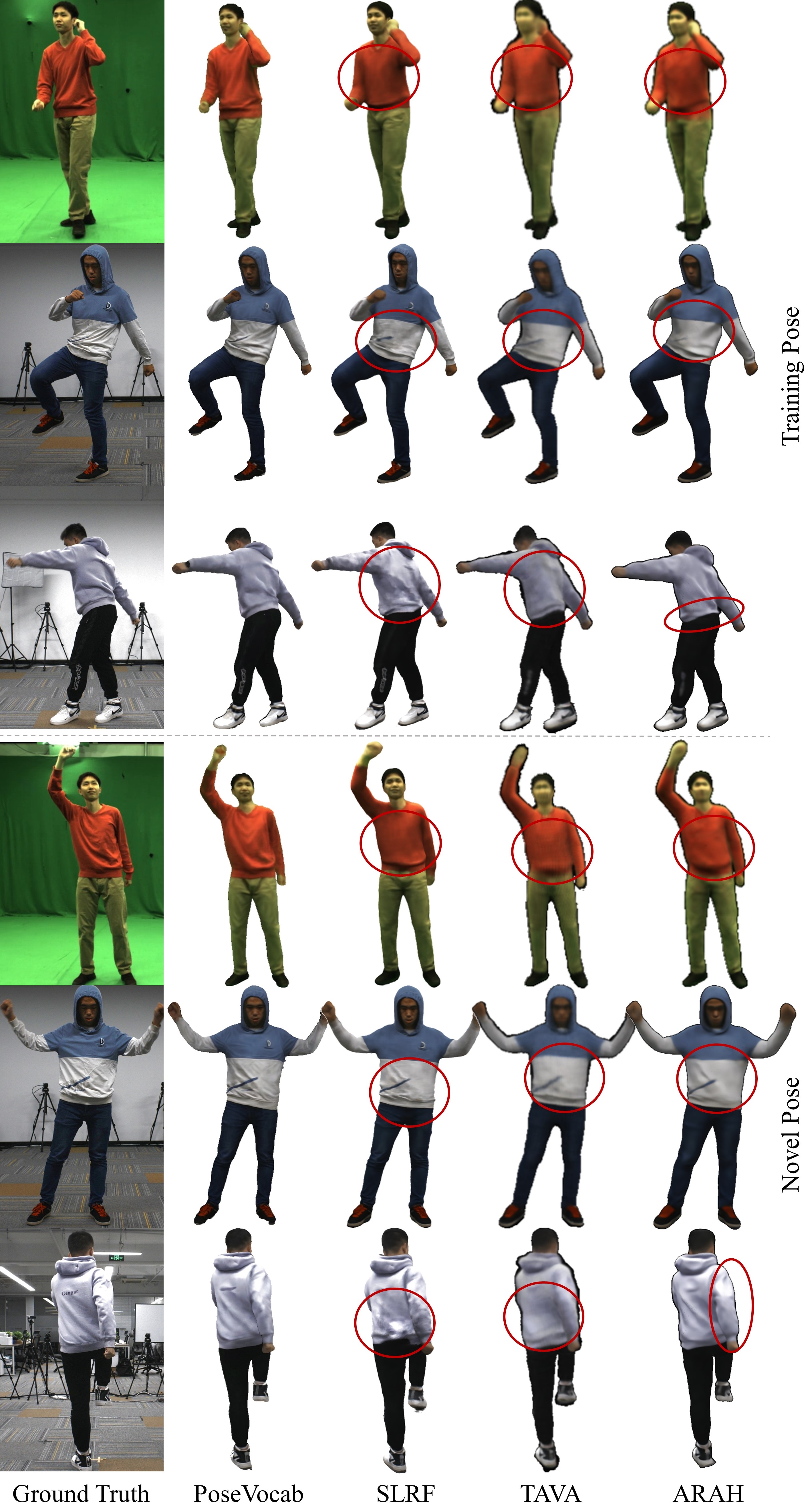}
    \caption{\textbf{Qualitative comparison against SLRF \cite{zheng2022structured}, TAVA \cite{li2022tava} and ARAH \cite{wang2022arah} on DeepCap and THuman4.0 datasets.} We show ground-truth images and animated avatars by PoseVocab (our method), SLRF, TAVA and ARAH under both training and novel poses, respectively.
    }
    \label{fig:cmp_slrf}
\end{minipage}
\hfill
\begin{minipage}{0.48\linewidth}
    \centering
    \includegraphics[width=\linewidth]{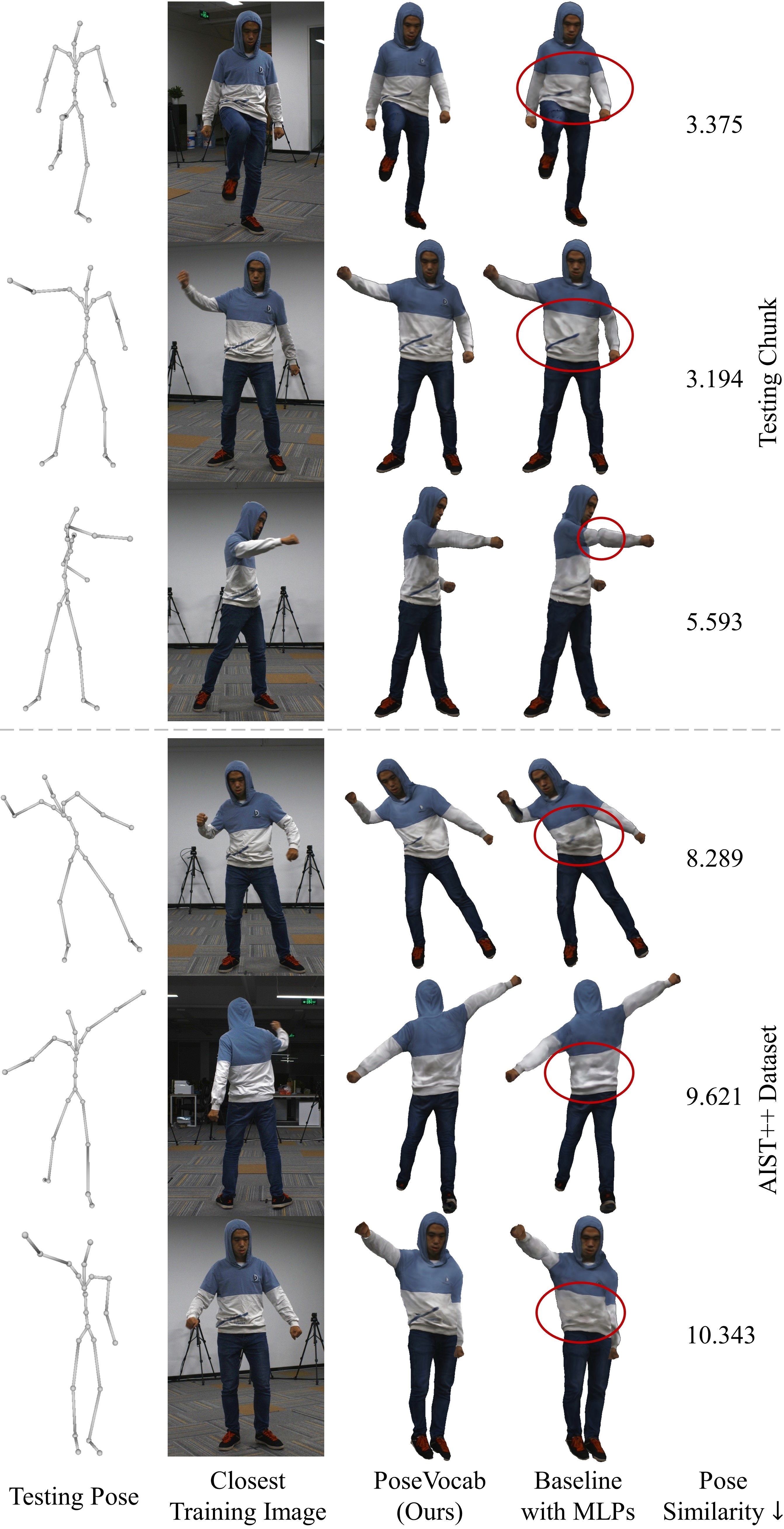}
    \caption{{
    \textbf{Evaluation on pose generalization.} 
    We compare our method with the baseline method that replaces PoseVocab with MLPs under novel testing poses.
    The testing poses are from the testing chunk and AIST++ dataset \cite{li2021ai} that contains lots of fancy poses.
    The closest training image is selected in the training dataset by pose similarity and view similarity.
    The pose similarity is computed as a face-area-weighted average distance (unit: cm) between SMPL vertices driven by the testing and training poses.
    }}
    \label{fig:eval_pose_generalization}
\end{minipage}
\end{figure*}

%% file: main.bbl

\begin{thebibliography}{77}

%%% the Web address.

\ifx \showCODEN    \undefined \def \showCODEN     #1{\unskip}     \fi
\ifx \showDOI      \undefined \def \showDOI       #1{#1}\fi
\ifx \showISBNx    \undefined \def \showISBNx     #1{\unskip}     \fi
\ifx \showISBNxiii \undefined \def \showISBNxiii  #1{\unskip}     \fi
\ifx \showISSN     \undefined \def \showISSN      #1{\unskip}     \fi
\ifx \showLCCN     \undefined \def \showLCCN      #1{\unskip}     \fi
\ifx \shownote     \undefined \def \shownote      #1{#1}          \fi
\ifx \showarticletitle \undefined \def \showarticletitle #1{#1}   \fi
\ifx \showURL      \undefined \def \showURL       {\relax}        \fi
% invisible to TeX
\providecommand\bibfield[2]{#2}
\providecommand\bibinfo[2]{#2}
\providecommand\natexlab[1]{#1}
\providecommand\showeprint[2][]{arXiv:#2}

\bibitem[Alldieck et~al\mbox{.}(2018)]%
        {alldieck2018video}
\bibfield{author}{\bibinfo{person}{Thiemo Alldieck}, \bibinfo{person}{Marcus
  Magnor}, \bibinfo{person}{Weipeng Xu}, \bibinfo{person}{Christian Theobalt},
  {and} \bibinfo{person}{Gerard Pons-Moll}.} \bibinfo{year}{2018}\natexlab{}.
\newblock \showarticletitle{Video based reconstruction of 3d people models}. In
  \bibinfo{booktitle}{\emph{CVPR}}. \bibinfo{pages}{8387--8397}.
\newblock


\bibitem[Alldieck et~al\mbox{.}(2019)]%
        {alldieck2019tex2shape}
\bibfield{author}{\bibinfo{person}{Thiemo Alldieck}, \bibinfo{person}{Gerard
  Pons-Moll}, \bibinfo{person}{Christian Theobalt}, {and}
  \bibinfo{person}{Marcus Magnor}.} \bibinfo{year}{2019}\natexlab{}.
\newblock \showarticletitle{Tex2shape: Detailed full human body geometry from a
  single image}. In \bibinfo{booktitle}{\emph{ICCV}}.
  \bibinfo{pages}{2293--2303}.
\newblock


\bibitem[Bagautdinov et~al\mbox{.}(2021)]%
        {bagautdinov2021driving}
\bibfield{author}{\bibinfo{person}{Timur Bagautdinov},
  \bibinfo{person}{Chenglei Wu}, \bibinfo{person}{Tomas Simon},
  \bibinfo{person}{Fabian Prada}, \bibinfo{person}{Takaaki Shiratori},
  \bibinfo{person}{Shih-En Wei}, \bibinfo{person}{Weipeng Xu},
  \bibinfo{person}{Yaser Sheikh}, {and} \bibinfo{person}{Jason Saragih}.}
  \bibinfo{year}{2021}\natexlab{}.
\newblock \showarticletitle{Driving-signal aware full-body avatars}.
\newblock \bibinfo{journal}{\emph{TOG}} \bibinfo{volume}{40},
  \bibinfo{number}{4} (\bibinfo{year}{2021}), \bibinfo{pages}{1--17}.
\newblock


\bibitem[Burov et~al\mbox{.}(2021)]%
        {burov2021dynamic}
\bibfield{author}{\bibinfo{person}{Andrei Burov}, \bibinfo{person}{Matthias
  Nie{\ss}ner}, {and} \bibinfo{person}{Justus Thies}.}
  \bibinfo{year}{2021}\natexlab{}.
\newblock \showarticletitle{Dynamic surface function networks for clothed human
  bodies}. In \bibinfo{booktitle}{\emph{ICCV}}. \bibinfo{pages}{10754--10764}.
\newblock


\bibitem[Chan et~al\mbox{.}(2022)]%
        {chan2022efficient}
\bibfield{author}{\bibinfo{person}{Eric~R Chan}, \bibinfo{person}{Connor~Z
  Lin}, \bibinfo{person}{Matthew~A Chan}, \bibinfo{person}{Koki Nagano},
  \bibinfo{person}{Boxiao Pan}, \bibinfo{person}{Shalini De~Mello},
  \bibinfo{person}{Orazio Gallo}, \bibinfo{person}{Leonidas~J Guibas},
  \bibinfo{person}{Jonathan Tremblay}, \bibinfo{person}{Sameh Khamis},
  {et~al\mbox{.}}} \bibinfo{year}{2022}\natexlab{}.
\newblock \showarticletitle{Efficient geometry-aware 3D generative adversarial
  networks}. In \bibinfo{booktitle}{\emph{CVPR}}.
  \bibinfo{pages}{16123--16133}.
\newblock


\bibitem[Chen et~al\mbox{.}(2022)]%
        {Chen2022TensoRF}
\bibfield{author}{\bibinfo{person}{Anpei Chen}, \bibinfo{person}{Zexiang Xu},
  \bibinfo{person}{Andreas Geiger}, \bibinfo{person}{Jingyi Yu}, {and}
  \bibinfo{person}{Hao Su}.} \bibinfo{year}{2022}\natexlab{}.
\newblock \showarticletitle{TensoRF: Tensorial Radiance Fields}. In
  \bibinfo{booktitle}{\emph{ECCV}}.
\newblock


\bibitem[Chen et~al\mbox{.}(2021)]%
        {chen2021snarf}
\bibfield{author}{\bibinfo{person}{Xu Chen}, \bibinfo{person}{Yufeng Zheng},
  \bibinfo{person}{Michael~J Black}, \bibinfo{person}{Otmar Hilliges}, {and}
  \bibinfo{person}{Andreas Geiger}.} \bibinfo{year}{2021}\natexlab{}.
\newblock \showarticletitle{SNARF: Differentiable forward skinning for
  animating non-rigid neural implicit shapes}. In
  \bibinfo{booktitle}{\emph{ICCV}}. \bibinfo{pages}{11594--11604}.
\newblock


\bibitem[Corona et~al\mbox{.}(2022)]%
        {corona2022lisa}
\bibfield{author}{\bibinfo{person}{Enric Corona}, \bibinfo{person}{Tomas
  Hodan}, \bibinfo{person}{Minh Vo}, \bibinfo{person}{Francesc Moreno-Noguer},
  \bibinfo{person}{Chris Sweeney}, \bibinfo{person}{Richard Newcombe}, {and}
  \bibinfo{person}{Lingni Ma}.} \bibinfo{year}{2022}\natexlab{}.
\newblock \showarticletitle{LISA: Learning implicit shape and appearance of
  hands}. In \bibinfo{booktitle}{\emph{CVPR}}. \bibinfo{pages}{20533--20543}.
\newblock


\bibitem[Deng et~al\mbox{.}(2020)]%
        {deng2020nasa}
\bibfield{author}{\bibinfo{person}{Boyang Deng}, \bibinfo{person}{John~P
  Lewis}, \bibinfo{person}{Timothy Jeruzalski}, \bibinfo{person}{Gerard
  Pons-Moll}, \bibinfo{person}{Geoffrey Hinton}, \bibinfo{person}{Mohammad
  Norouzi}, {and} \bibinfo{person}{Andrea Tagliasacchi}.}
  \bibinfo{year}{2020}\natexlab{}.
\newblock \showarticletitle{NASA neural articulated shape approximation}. In
  \bibinfo{booktitle}{\emph{ECCV}}. Springer, \bibinfo{pages}{612--628}.
\newblock


\bibitem[Dong et~al\mbox{.}(2022a)]%
        {dong2022totalselfscan}
\bibfield{author}{\bibinfo{person}{Junting Dong}, \bibinfo{person}{Qi Fang},
  \bibinfo{person}{Yudong Guo}, \bibinfo{person}{Sida Peng},
  \bibinfo{person}{Qing Shuai}, \bibinfo{person}{Xiaowei Zhou}, {and}
  \bibinfo{person}{Hujun Bao}.} \bibinfo{year}{2022}\natexlab{a}.
\newblock \showarticletitle{TotalSelfScan: Learning Full-body Avatars from
  Self-Portrait Videos of Faces, Hands, and Bodies}. In
  \bibinfo{booktitle}{\emph{NeurIPS}}.
\newblock


\bibitem[Dong et~al\mbox{.}(2022b)]%
        {dong2022pina}
\bibfield{author}{\bibinfo{person}{Zijian Dong}, \bibinfo{person}{Chen Guo},
  \bibinfo{person}{Jie Song}, \bibinfo{person}{Xu Chen},
  \bibinfo{person}{Andreas Geiger}, {and} \bibinfo{person}{Otmar Hilliges}.}
  \bibinfo{year}{2022}\natexlab{b}.
\newblock \showarticletitle{PINA: Learning a Personalized Implicit Neural
  Avatar from a Single RGB-D Video Sequence}. In
  \bibinfo{booktitle}{\emph{CVPR}}.
\newblock


\bibitem[Feng et~al\mbox{.}(2022)]%
        {Feng2022scarf}
\bibfield{author}{\bibinfo{person}{Yao Feng}, \bibinfo{person}{Jinlong Yang},
  \bibinfo{person}{Marc Pollefeys}, \bibinfo{person}{Michael~J. Black}, {and}
  \bibinfo{person}{Timo Bolkart}.} \bibinfo{year}{2022}\natexlab{}.
\newblock \showarticletitle{Capturing and Animation of Body and Clothing from
  Monocular Video}. In \bibinfo{booktitle}{\emph{SIGGRAPH Asia 2022 Conference
  Proceedings}} (Daegu, Republic of Korea) \emph{(\bibinfo{series}{SA '22})}.
  Article \bibinfo{articleno}{45}, \bibinfo{numpages}{9}~pages.
\newblock


\bibitem[Fridovich-Keil et~al\mbox{.}(2022)]%
        {fridovich2022plenoxels}
\bibfield{author}{\bibinfo{person}{Sara Fridovich-Keil}, \bibinfo{person}{Alex
  Yu}, \bibinfo{person}{Matthew Tancik}, \bibinfo{person}{Qinhong Chen},
  \bibinfo{person}{Benjamin Recht}, {and} \bibinfo{person}{Angjoo Kanazawa}.}
  \bibinfo{year}{2022}\natexlab{}.
\newblock \showarticletitle{Plenoxels: Radiance Fields Without Neural
  Networks}. In \bibinfo{booktitle}{\emph{CVPR}}. \bibinfo{pages}{5501--5510}.
\newblock


\bibitem[Gao et~al\mbox{.}(2022)]%
        {gao2022reconstructing}
\bibfield{author}{\bibinfo{person}{Xuan Gao}, \bibinfo{person}{Chenglai Zhong},
  \bibinfo{person}{Jun Xiang}, \bibinfo{person}{Yang Hong},
  \bibinfo{person}{Yudong Guo}, {and} \bibinfo{person}{Juyong Zhang}.}
  \bibinfo{year}{2022}\natexlab{}.
\newblock \showarticletitle{Reconstructing personalized semantic facial nerf
  models from monocular video}.
\newblock \bibinfo{journal}{\emph{ACM Transactions on Graphics (TOG)}}
  \bibinfo{volume}{41}, \bibinfo{number}{6} (\bibinfo{year}{2022}),
  \bibinfo{pages}{1--12}.
\newblock


\bibitem[Gropp et~al\mbox{.}(2020)]%
        {gropp2020implicit}
\bibfield{author}{\bibinfo{person}{Amos Gropp}, \bibinfo{person}{Lior Yariv},
  \bibinfo{person}{Niv Haim}, \bibinfo{person}{Matan Atzmon}, {and}
  \bibinfo{person}{Yaron Lipman}.} \bibinfo{year}{2020}\natexlab{}.
\newblock \showarticletitle{Implicit Geometric Regularization for Learning
  Shapes}. In \bibinfo{booktitle}{\emph{ICML}}. PMLR,
  \bibinfo{pages}{3789--3799}.
\newblock


\bibitem[Guo et~al\mbox{.}(2022)]%
        {guo2022generating}
\bibfield{author}{\bibinfo{person}{Chuan Guo}, \bibinfo{person}{Shihao Zou},
  \bibinfo{person}{Xinxin Zuo}, \bibinfo{person}{Sen Wang},
  \bibinfo{person}{Wei Ji}, \bibinfo{person}{Xingyu Li}, {and}
  \bibinfo{person}{Li Cheng}.} \bibinfo{year}{2022}\natexlab{}.
\newblock \showarticletitle{Generating diverse and natural 3d human motions
  from text}. In \bibinfo{booktitle}{\emph{CVPR}}. \bibinfo{pages}{5152--5161}.
\newblock


\bibitem[Habermann et~al\mbox{.}(2021)]%
        {habermann2021real}
\bibfield{author}{\bibinfo{person}{Marc Habermann}, \bibinfo{person}{Lingjie
  Liu}, \bibinfo{person}{Weipeng Xu}, \bibinfo{person}{Michael Zollhoefer},
  \bibinfo{person}{Gerard Pons-Moll}, {and} \bibinfo{person}{Christian
  Theobalt}.} \bibinfo{year}{2021}\natexlab{}.
\newblock \showarticletitle{Real-time deep dynamic characters}.
\newblock \bibinfo{journal}{\emph{TOG}} \bibinfo{volume}{40},
  \bibinfo{number}{4} (\bibinfo{year}{2021}), \bibinfo{pages}{1--16}.
\newblock


\bibitem[Habermann et~al\mbox{.}(2020)]%
        {habermann2020deepcap}
\bibfield{author}{\bibinfo{person}{Marc Habermann}, \bibinfo{person}{Weipeng
  Xu}, \bibinfo{person}{Michael Zollhofer}, \bibinfo{person}{Gerard Pons-Moll},
  {and} \bibinfo{person}{Christian Theobalt}.} \bibinfo{year}{2020}\natexlab{}.
\newblock \showarticletitle{Deepcap: Monocular human performance capture using
  weak supervision}. In \bibinfo{booktitle}{\emph{CVPR}}.
  \bibinfo{pages}{5052--5063}.
\newblock


\bibitem[Halimi et~al\mbox{.}(2022)]%
        {halimi2022pattern}
\bibfield{author}{\bibinfo{person}{Oshri Halimi}, \bibinfo{person}{Tuur
  Stuyck}, \bibinfo{person}{Donglai Xiang}, \bibinfo{person}{Timur
  Bagautdinov}, \bibinfo{person}{He Wen}, \bibinfo{person}{Ron Kimmel},
  \bibinfo{person}{Takaaki Shiratori}, \bibinfo{person}{Chenglei Wu},
  \bibinfo{person}{Yaser Sheikh}, {and} \bibinfo{person}{Fabian Prada}.}
  \bibinfo{year}{2022}\natexlab{}.
\newblock \showarticletitle{Pattern-Based Cloth Registration and Sparse-View
  Animation}.
\newblock \bibinfo{journal}{\emph{ACM Transactions on Graphics (TOG)}}
  \bibinfo{volume}{41}, \bibinfo{number}{6} (\bibinfo{year}{2022}),
  \bibinfo{pages}{1--17}.
\newblock


\bibitem[He et~al\mbox{.}(2021)]%
        {he2021arch++}
\bibfield{author}{\bibinfo{person}{Tong He}, \bibinfo{person}{Yuanlu Xu},
  \bibinfo{person}{Shunsuke Saito}, \bibinfo{person}{Stefano Soatto}, {and}
  \bibinfo{person}{Tony Tung}.} \bibinfo{year}{2021}\natexlab{}.
\newblock \showarticletitle{ARCH++: Animation-ready clothed human
  reconstruction revisited}. In \bibinfo{booktitle}{\emph{ICCV}}.
  \bibinfo{pages}{11046--11056}.
\newblock


\bibitem[Heusel et~al\mbox{.}(2017)]%
        {heusel2017gans}
\bibfield{author}{\bibinfo{person}{Martin Heusel}, \bibinfo{person}{Hubert
  Ramsauer}, \bibinfo{person}{Thomas Unterthiner}, \bibinfo{person}{Bernhard
  Nessler}, {and} \bibinfo{person}{Sepp Hochreiter}.}
  \bibinfo{year}{2017}\natexlab{}.
\newblock \showarticletitle{Gans trained by a two time-scale update rule
  converge to a local nash equilibrium}.
\newblock \bibinfo{journal}{\emph{NeurIPS}}  \bibinfo{volume}{30}
  (\bibinfo{year}{2017}).
\newblock


\bibitem[Huang et~al\mbox{.}(2022)]%
        {huang2022one}
\bibfield{author}{\bibinfo{person}{Yangyi Huang}, \bibinfo{person}{Hongwei Yi},
  \bibinfo{person}{Weiyang Liu}, \bibinfo{person}{Haofan Wang},
  \bibinfo{person}{Boxi Wu}, \bibinfo{person}{Wenxiao Wang},
  \bibinfo{person}{Binbin Lin}, \bibinfo{person}{Debing Zhang}, {and}
  \bibinfo{person}{Deng Cai}.} \bibinfo{year}{2022}\natexlab{}.
\newblock \showarticletitle{One-shot Implicit Animatable Avatars with
  Model-based Priors}.
\newblock \bibinfo{journal}{\emph{arXiv preprint arXiv:2212.02469}}
  (\bibinfo{year}{2022}).
\newblock


\bibitem[Huang et~al\mbox{.}(2020)]%
        {huang2020arch}
\bibfield{author}{\bibinfo{person}{Zeng Huang}, \bibinfo{person}{Yuanlu Xu},
  \bibinfo{person}{Christoph Lassner}, \bibinfo{person}{Hao Li}, {and}
  \bibinfo{person}{Tony Tung}.} \bibinfo{year}{2020}\natexlab{}.
\newblock \showarticletitle{Arch: Animatable reconstruction of clothed humans}.
  In \bibinfo{booktitle}{\emph{CVPR}}. \bibinfo{pages}{3093--3102}.
\newblock


\bibitem[Huynh(2009)]%
        {huynh2009metrics}
\bibfield{author}{\bibinfo{person}{Du~Q Huynh}.}
  \bibinfo{year}{2009}\natexlab{}.
\newblock \showarticletitle{Metrics for 3D rotations: Comparison and analysis}.
\newblock \bibinfo{journal}{\emph{Journal of Mathematical Imaging and Vision}}
  \bibinfo{volume}{35}, \bibinfo{number}{2} (\bibinfo{year}{2009}),
  \bibinfo{pages}{155--164}.
\newblock


\bibitem[Jiang et~al\mbox{.}(2022b)]%
        {jiang2022selfrecon}
\bibfield{author}{\bibinfo{person}{Boyi Jiang}, \bibinfo{person}{Yang Hong},
  \bibinfo{person}{Hujun Bao}, {and} \bibinfo{person}{Juyong Zhang}.}
  \bibinfo{year}{2022}\natexlab{b}.
\newblock \showarticletitle{SelfRecon: Self Reconstruction Your Digital Avatar
  from Monocular Video}. In \bibinfo{booktitle}{\emph{CVPR}}.
  \bibinfo{pages}{5605--5615}.
\newblock


\bibitem[Jiang et~al\mbox{.}(2022a)]%
        {jiang2022instantavatar}
\bibfield{author}{\bibinfo{person}{Tianjian Jiang}, \bibinfo{person}{Xu Chen},
  \bibinfo{person}{Jie Song}, {and} \bibinfo{person}{Otmar Hilliges}.}
  \bibinfo{year}{2022}\natexlab{a}.
\newblock \showarticletitle{InstantAvatar: Learning Avatars from Monocular
  Video in 60 Seconds}.
\newblock \bibinfo{journal}{\emph{arXiv preprint arXiv:2212.10550}}
  (\bibinfo{year}{2022}).
\newblock


\bibitem[Jiang et~al\mbox{.}(2022c)]%
        {jiang2022neuman}
\bibfield{author}{\bibinfo{person}{Wei Jiang}, \bibinfo{person}{Kwang~Moo Yi},
  \bibinfo{person}{Golnoosh Samei}, \bibinfo{person}{Oncel Tuzel}, {and}
  \bibinfo{person}{Anurag Ranjan}.} \bibinfo{year}{2022}\natexlab{c}.
\newblock \showarticletitle{Neuman: Neural human radiance field from a single
  video}. In \bibinfo{booktitle}{\emph{ECCV}}. Springer,
  \bibinfo{pages}{402--418}.
\newblock


\bibitem[Kim et~al\mbox{.}(2022)]%
        {kim2022laplacianfusion}
\bibfield{author}{\bibinfo{person}{Hyomin Kim}, \bibinfo{person}{Hyeonseo Nam},
  \bibinfo{person}{Jungeon Kim}, \bibinfo{person}{Jaesik Park}, {and}
  \bibinfo{person}{Seungyong Lee}.} \bibinfo{year}{2022}\natexlab{}.
\newblock \showarticletitle{LaplacianFusion: Detailed 3D Clothed-Human Body
  Reconstruction}.
\newblock \bibinfo{journal}{\emph{ACM Transactions on Graphics (TOG)}}
  \bibinfo{volume}{41}, \bibinfo{number}{6} (\bibinfo{year}{2022}),
  \bibinfo{pages}{1--14}.
\newblock


\bibitem[Kingma and Ba(2015)]%
        {adam}
\bibfield{author}{\bibinfo{person}{Diederik~P Kingma} {and}
  \bibinfo{person}{Jimmy Ba}.} \bibinfo{year}{2015}\natexlab{}.
\newblock \showarticletitle{Adam: A method for stochastic optimization}. In
  \bibinfo{booktitle}{\emph{ICLR}}.
\newblock


\bibitem[Kipf and Welling(2016)]%
        {kipf2016semi}
\bibfield{author}{\bibinfo{person}{Thomas~N Kipf} {and} \bibinfo{person}{Max
  Welling}.} \bibinfo{year}{2016}\natexlab{}.
\newblock \showarticletitle{Semi-supervised classification with graph
  convolutional networks}.
\newblock \bibinfo{journal}{\emph{arXiv preprint arXiv:1609.02907}}
  (\bibinfo{year}{2016}).
\newblock


\bibitem[Li et~al\mbox{.}(2022a)]%
        {li2022tava}
\bibfield{author}{\bibinfo{person}{Ruilong Li}, \bibinfo{person}{Julian Tanke},
  \bibinfo{person}{Minh Vo}, \bibinfo{person}{Michael Zollh{\"o}fer},
  \bibinfo{person}{J{\"u}rgen Gall}, \bibinfo{person}{Angjoo Kanazawa}, {and}
  \bibinfo{person}{Christoph Lassner}.} \bibinfo{year}{2022}\natexlab{a}.
\newblock \showarticletitle{Tava: Template-free animatable volumetric actors}.
  In \bibinfo{booktitle}{\emph{ECCV}}. Springer, \bibinfo{pages}{419--436}.
\newblock


\bibitem[Li et~al\mbox{.}(2021)]%
        {li2021ai}
\bibfield{author}{\bibinfo{person}{Ruilong Li}, \bibinfo{person}{Shan Yang},
  \bibinfo{person}{David~A Ross}, {and} \bibinfo{person}{Angjoo Kanazawa}.}
  \bibinfo{year}{2021}\natexlab{}.
\newblock \showarticletitle{Ai choreographer: Music conditioned 3d dance
  generation with aist++}. In \bibinfo{booktitle}{\emph{ICCV}}.
  \bibinfo{pages}{13401--13412}.
\newblock


\bibitem[Li et~al\mbox{.}(2022b)]%
        {li2022avatarcap}
\bibfield{author}{\bibinfo{person}{Zhe Li}, \bibinfo{person}{Zerong Zheng},
  \bibinfo{person}{Hongwen Zhang}, \bibinfo{person}{Chaonan Ji}, {and}
  \bibinfo{person}{Yebin Liu}.} \bibinfo{year}{2022}\natexlab{b}.
\newblock \showarticletitle{AvatarCap: Animatable Avatar Conditioned Monocular
  Human Volumetric Capture}. In \bibinfo{booktitle}{\emph{ECCV}}. Springer,
  \bibinfo{pages}{322--341}.
\newblock


\bibitem[Lin et~al\mbox{.}(2022)]%
        {lin2022learning}
\bibfield{author}{\bibinfo{person}{Siyou Lin}, \bibinfo{person}{Hongwen Zhang},
  \bibinfo{person}{Zerong Zheng}, \bibinfo{person}{Ruizhi Shao}, {and}
  \bibinfo{person}{Yebin Liu}.} \bibinfo{year}{2022}\natexlab{}.
\newblock \showarticletitle{Learning implicit templates for point-based clothed
  human modeling}. In \bibinfo{booktitle}{\emph{ECCV}}. Springer,
  \bibinfo{pages}{210--228}.
\newblock


\bibitem[Liu et~al\mbox{.}(2021)]%
        {liu2021neural}
\bibfield{author}{\bibinfo{person}{Lingjie Liu}, \bibinfo{person}{Marc
  Habermann}, \bibinfo{person}{Viktor Rudnev}, \bibinfo{person}{Kripasindhu
  Sarkar}, \bibinfo{person}{Jiatao Gu}, {and} \bibinfo{person}{Christian
  Theobalt}.} \bibinfo{year}{2021}\natexlab{}.
\newblock \showarticletitle{Neural actor: Neural free-view synthesis of human
  actors with pose control}.
\newblock \bibinfo{journal}{\emph{TOG}} \bibinfo{volume}{40},
  \bibinfo{number}{6} (\bibinfo{year}{2021}), \bibinfo{pages}{1--16}.
\newblock


\bibitem[Loper et~al\mbox{.}(2015)]%
        {loper2015smpl}
\bibfield{author}{\bibinfo{person}{Matthew Loper}, \bibinfo{person}{Naureen
  Mahmood}, \bibinfo{person}{Javier Romero}, \bibinfo{person}{Gerard
  Pons-Moll}, {and} \bibinfo{person}{Michael~J Black}.}
  \bibinfo{year}{2015}\natexlab{}.
\newblock \showarticletitle{SMPL: A skinned multi-person linear model}.
\newblock \bibinfo{journal}{\emph{TOG}} \bibinfo{volume}{34},
  \bibinfo{number}{6} (\bibinfo{year}{2015}), \bibinfo{pages}{1--16}.
\newblock


\bibitem[Lorensen and Cline(1987)]%
        {lorensen1987marching}
\bibfield{author}{\bibinfo{person}{William~E Lorensen} {and}
  \bibinfo{person}{Harvey~E Cline}.} \bibinfo{year}{1987}\natexlab{}.
\newblock \showarticletitle{Marching cubes: A high resolution 3D surface
  construction algorithm}.
\newblock \bibinfo{journal}{\emph{TOG}} \bibinfo{volume}{21},
  \bibinfo{number}{4} (\bibinfo{year}{1987}), \bibinfo{pages}{163--169}.
\newblock


\bibitem[Ma et~al\mbox{.}(2021a)]%
        {ma2021scale}
\bibfield{author}{\bibinfo{person}{Qianli Ma}, \bibinfo{person}{Shunsuke
  Saito}, \bibinfo{person}{Jinlong Yang}, \bibinfo{person}{Siyu Tang}, {and}
  \bibinfo{person}{Michael~J Black}.} \bibinfo{year}{2021}\natexlab{a}.
\newblock \showarticletitle{SCALE: Modeling clothed humans with a surface codec
  of articulated local elements}. In \bibinfo{booktitle}{\emph{CVPR}}.
  \bibinfo{pages}{16082--16093}.
\newblock


\bibitem[Ma et~al\mbox{.}(2021b)]%
        {ma2021power}
\bibfield{author}{\bibinfo{person}{Qianli Ma}, \bibinfo{person}{Jinlong Yang},
  \bibinfo{person}{Siyu Tang}, {and} \bibinfo{person}{Michael~J Black}.}
  \bibinfo{year}{2021}\natexlab{b}.
\newblock \showarticletitle{The power of points for modeling humans in
  clothing}. In \bibinfo{booktitle}{\emph{ICCV}}.
  \bibinfo{pages}{10974--10984}.
\newblock


\bibitem[Mihajlovic et~al\mbox{.}(2022)]%
        {mihajlovic2022coap}
\bibfield{author}{\bibinfo{person}{Marko Mihajlovic}, \bibinfo{person}{Shunsuke
  Saito}, \bibinfo{person}{Aayush Bansal}, \bibinfo{person}{Michael
  Zollhoefer}, {and} \bibinfo{person}{Siyu Tang}.}
  \bibinfo{year}{2022}\natexlab{}.
\newblock \showarticletitle{COAP: Compositional articulated occupancy of
  people}. In \bibinfo{booktitle}{\emph{CVPR}}. \bibinfo{pages}{13201--13210}.
\newblock


\bibitem[Mihajlovic et~al\mbox{.}(2021)]%
        {mihajlovic2021leap}
\bibfield{author}{\bibinfo{person}{Marko Mihajlovic}, \bibinfo{person}{Yan
  Zhang}, \bibinfo{person}{Michael~J Black}, {and} \bibinfo{person}{Siyu
  Tang}.} \bibinfo{year}{2021}\natexlab{}.
\newblock \showarticletitle{LEAP: Learning articulated occupancy of people}. In
  \bibinfo{booktitle}{\emph{CVPR}}. \bibinfo{pages}{10461--10471}.
\newblock


\bibitem[Mikolov et~al\mbox{.}(2013a)]%
        {mikolov2013efficient}
\bibfield{author}{\bibinfo{person}{Tomas Mikolov}, \bibinfo{person}{Kai Chen},
  \bibinfo{person}{Greg Corrado}, {and} \bibinfo{person}{Jeffrey Dean}.}
  \bibinfo{year}{2013}\natexlab{a}.
\newblock \showarticletitle{Efficient estimation of word representations in
  vector space}.
\newblock \bibinfo{journal}{\emph{arXiv preprint arXiv:1301.3781}}
  (\bibinfo{year}{2013}).
\newblock


\bibitem[Mikolov et~al\mbox{.}(2013b)]%
        {mikolov2013distributed}
\bibfield{author}{\bibinfo{person}{Tomas Mikolov}, \bibinfo{person}{Ilya
  Sutskever}, \bibinfo{person}{Kai Chen}, \bibinfo{person}{Greg~S Corrado},
  {and} \bibinfo{person}{Jeff Dean}.} \bibinfo{year}{2013}\natexlab{b}.
\newblock \showarticletitle{Distributed representations of words and phrases
  and their compositionality}.
\newblock \bibinfo{journal}{\emph{NeurIPS}}  \bibinfo{volume}{26}
  (\bibinfo{year}{2013}).
\newblock


\bibitem[Mildenhall et~al\mbox{.}(2020)]%
        {mildenhall2020nerf}
\bibfield{author}{\bibinfo{person}{Ben Mildenhall}, \bibinfo{person}{Pratul~P
  Srinivasan}, \bibinfo{person}{Matthew Tancik}, \bibinfo{person}{Jonathan~T
  Barron}, \bibinfo{person}{Ravi Ramamoorthi}, {and} \bibinfo{person}{Ren Ng}.}
  \bibinfo{year}{2020}\natexlab{}.
\newblock \showarticletitle{Nerf: Representing scenes as neural radiance fields
  for view synthesis}. In \bibinfo{booktitle}{\emph{ECCV}}. Springer,
  \bibinfo{pages}{405--421}.
\newblock


\bibitem[M\"{u}ller et~al\mbox{.}(2022)]%
        {instantngp}
\bibfield{author}{\bibinfo{person}{Thomas M\"{u}ller}, \bibinfo{person}{Alex
  Evans}, \bibinfo{person}{Christoph Schied}, {and} \bibinfo{person}{Alexander
  Keller}.} \bibinfo{year}{2022}\natexlab{}.
\newblock \showarticletitle{Instant Neural Graphics Primitives with a
  Multiresolution Hash Encoding}.
\newblock \bibinfo{journal}{\emph{ACM Transactions on Graphics (TOG)}}
  \bibinfo{volume}{41}, \bibinfo{number}{4}, Article \bibinfo{articleno}{102}
  (\bibinfo{date}{jul} \bibinfo{year}{2022}), \bibinfo{numpages}{15}~pages.
\newblock
\showISSN{0730-0301}
\urldef\tempurl%
\url{https://doi.org/10.1145/3528223.3530127}
\showDOI{\tempurl}


\bibitem[Park et~al\mbox{.}(2019)]%
        {park2019deepsdf}
\bibfield{author}{\bibinfo{person}{Jeong~Joon Park}, \bibinfo{person}{Peter
  Florence}, \bibinfo{person}{Julian Straub}, \bibinfo{person}{Richard
  Newcombe}, {and} \bibinfo{person}{Steven Lovegrove}.}
  \bibinfo{year}{2019}\natexlab{}.
\newblock \showarticletitle{Deepsdf: Learning continuous signed distance
  functions for shape representation}. In \bibinfo{booktitle}{\emph{CVPR}}.
  \bibinfo{pages}{165--174}.
\newblock


\bibitem[Paszke et~al\mbox{.}(2019)]%
        {paszke2019pytorch}
\bibfield{author}{\bibinfo{person}{Adam Paszke}, \bibinfo{person}{Sam Gross},
  \bibinfo{person}{Francisco Massa}, \bibinfo{person}{Adam Lerer},
  \bibinfo{person}{James Bradbury}, \bibinfo{person}{Gregory Chanan},
  \bibinfo{person}{Trevor Killeen}, \bibinfo{person}{Zeming Lin},
  \bibinfo{person}{Natalia Gimelshein}, \bibinfo{person}{Luca Antiga},
  {et~al\mbox{.}}} \bibinfo{year}{2019}\natexlab{}.
\newblock \showarticletitle{Pytorch: An imperative style, high-performance deep
  learning library}.
\newblock \bibinfo{journal}{\emph{NeurIPS}}  \bibinfo{volume}{32}
  (\bibinfo{year}{2019}).
\newblock


\bibitem[Pavlakos et~al\mbox{.}(2019)]%
        {SMPL-X:2019}
\bibfield{author}{\bibinfo{person}{Georgios Pavlakos},
  \bibinfo{person}{Vasileios Choutas}, \bibinfo{person}{Nima Ghorbani},
  \bibinfo{person}{Timo Bolkart}, \bibinfo{person}{Ahmed A.~A. Osman},
  \bibinfo{person}{Dimitrios Tzionas}, {and} \bibinfo{person}{Michael~J.
  Black}.} \bibinfo{year}{2019}\natexlab{}.
\newblock \showarticletitle{Expressive Body Capture: 3D Hands, Face, and Body
  from a Single Image}. In \bibinfo{booktitle}{\emph{CVPR}}.
\newblock


\bibitem[Peng et~al\mbox{.}(2022a)]%
        {peng2022selfnerf}
\bibfield{author}{\bibinfo{person}{Bo Peng}, \bibinfo{person}{Jun Hu},
  \bibinfo{person}{Jingtao Zhou}, {and} \bibinfo{person}{Juyong Zhang}.}
  \bibinfo{year}{2022}\natexlab{a}.
\newblock \showarticletitle{SelfNeRF: Fast Training NeRF for Human from
  Monocular Self-rotating Video}.
\newblock \bibinfo{journal}{\emph{arXiv preprint arXiv:2210.01651}}
  (\bibinfo{year}{2022}).
\newblock


\bibitem[Peng et~al\mbox{.}(2021a)]%
        {peng2021animatable}
\bibfield{author}{\bibinfo{person}{Sida Peng}, \bibinfo{person}{Junting Dong},
  \bibinfo{person}{Qianqian Wang}, \bibinfo{person}{Shangzhan Zhang},
  \bibinfo{person}{Qing Shuai}, \bibinfo{person}{Xiaowei Zhou}, {and}
  \bibinfo{person}{Hujun Bao}.} \bibinfo{year}{2021}\natexlab{a}.
\newblock \showarticletitle{Animatable neural radiance fields for modeling
  dynamic human bodies}. In \bibinfo{booktitle}{\emph{ICCV}}.
  \bibinfo{pages}{14314--14323}.
\newblock


\bibitem[Peng et~al\mbox{.}(2022b)]%
        {peng2022animatable}
\bibfield{author}{\bibinfo{person}{Sida Peng}, \bibinfo{person}{Shangzhan
  Zhang}, \bibinfo{person}{Zhen Xu}, \bibinfo{person}{Chen Geng},
  \bibinfo{person}{Boyi Jiang}, \bibinfo{person}{Hujun Bao}, {and}
  \bibinfo{person}{Xiaowei Zhou}.} \bibinfo{year}{2022}\natexlab{b}.
\newblock \showarticletitle{Animatable Neural Implicit Surfaces for Creating
  Avatars from Videos}.
\newblock \bibinfo{journal}{\emph{arXiv preprint arXiv:2203.08133}}
  (\bibinfo{year}{2022}).
\newblock


\bibitem[Peng et~al\mbox{.}(2021b)]%
        {peng2021neural}
\bibfield{author}{\bibinfo{person}{Sida Peng}, \bibinfo{person}{Yuanqing
  Zhang}, \bibinfo{person}{Yinghao Xu}, \bibinfo{person}{Qianqian Wang},
  \bibinfo{person}{Qing Shuai}, \bibinfo{person}{Hujun Bao}, {and}
  \bibinfo{person}{Xiaowei Zhou}.} \bibinfo{year}{2021}\natexlab{b}.
\newblock \showarticletitle{Neural body: Implicit neural representations with
  structured latent codes for novel view synthesis of dynamic humans}. In
  \bibinfo{booktitle}{\emph{CVPR}}. \bibinfo{pages}{9054--9063}.
\newblock


\bibitem[Remelli et~al\mbox{.}(2022)]%
        {remelli2022drivable}
\bibfield{author}{\bibinfo{person}{Edoardo Remelli}, \bibinfo{person}{Timur
  Bagautdinov}, \bibinfo{person}{Shunsuke Saito}, \bibinfo{person}{Chenglei
  Wu}, \bibinfo{person}{Tomas Simon}, \bibinfo{person}{Shih-En Wei},
  \bibinfo{person}{Kaiwen Guo}, \bibinfo{person}{Zhe Cao},
  \bibinfo{person}{Fabian Prada}, \bibinfo{person}{Jason Saragih},
  {et~al\mbox{.}}} \bibinfo{year}{2022}\natexlab{}.
\newblock \showarticletitle{Drivable volumetric avatars using texel-aligned
  features}. In \bibinfo{booktitle}{\emph{ACM SIGGRAPH 2022 Conference
  Proceedings}}. \bibinfo{pages}{1--9}.
\newblock


\bibitem[Saito et~al\mbox{.}(2021)]%
        {saito2021scanimate}
\bibfield{author}{\bibinfo{person}{Shunsuke Saito}, \bibinfo{person}{Jinlong
  Yang}, \bibinfo{person}{Qianli Ma}, {and} \bibinfo{person}{Michael~J Black}.}
  \bibinfo{year}{2021}\natexlab{}.
\newblock \showarticletitle{SCANimate: Weakly supervised learning of skinned
  clothed avatar networks}. In \bibinfo{booktitle}{\emph{CVPR}}.
  \bibinfo{pages}{2886--2897}.
\newblock


\bibitem[Su et~al\mbox{.}(2022)]%
        {su2022danbo}
\bibfield{author}{\bibinfo{person}{Shih-Yang Su}, \bibinfo{person}{Timur
  Bagautdinov}, {and} \bibinfo{person}{Helge Rhodin}.}
  \bibinfo{year}{2022}\natexlab{}.
\newblock \showarticletitle{Danbo: Disentangled articulated neural body
  representations via graph neural networks}. In
  \bibinfo{booktitle}{\emph{ECCV}}. Springer, \bibinfo{pages}{107--124}.
\newblock


\bibitem[Su et~al\mbox{.}(2021)]%
        {su2021a-nerf}
\bibfield{author}{\bibinfo{person}{Shih-Yang Su}, \bibinfo{person}{Frank Yu},
  \bibinfo{person}{Michael Zollh{\"o}fer}, {and} \bibinfo{person}{Helge
  Rhodin}.} \bibinfo{year}{2021}\natexlab{}.
\newblock \showarticletitle{A-nerf: Articulated neural radiance fields for
  learning human shape, appearance, and pose}.
\newblock \bibinfo{journal}{\emph{NeurIPS}}  \bibinfo{volume}{34}
  (\bibinfo{year}{2021}), \bibinfo{pages}{12278--12291}.
\newblock


\bibitem[Tancik et~al\mbox{.}(2020)]%
        {tancik2020fourier}
\bibfield{author}{\bibinfo{person}{Matthew Tancik}, \bibinfo{person}{Pratul
  Srinivasan}, \bibinfo{person}{Ben Mildenhall}, \bibinfo{person}{Sara
  Fridovich-Keil}, \bibinfo{person}{Nithin Raghavan}, \bibinfo{person}{Utkarsh
  Singhal}, \bibinfo{person}{Ravi Ramamoorthi}, \bibinfo{person}{Jonathan
  Barron}, {and} \bibinfo{person}{Ren Ng}.} \bibinfo{year}{2020}\natexlab{}.
\newblock \showarticletitle{Fourier features let networks learn high frequency
  functions in low dimensional domains}.
\newblock \bibinfo{journal}{\emph{NeurIPS}}  \bibinfo{volume}{33}
  (\bibinfo{year}{2020}), \bibinfo{pages}{7537--7547}.
\newblock


\bibitem[Te et~al\mbox{.}(2022)]%
        {te2022neural}
\bibfield{author}{\bibinfo{person}{Gusi Te}, \bibinfo{person}{Xiu Li},
  \bibinfo{person}{Xiao Li}, \bibinfo{person}{Jinglu Wang},
  \bibinfo{person}{Wei Hu}, {and} \bibinfo{person}{Yan Lu}.}
  \bibinfo{year}{2022}\natexlab{}.
\newblock \showarticletitle{Neural Capture of Animatable 3D Human from
  Monocular Video}. In \bibinfo{booktitle}{\emph{ECCV}}. Springer,
  \bibinfo{pages}{275--291}.
\newblock


\bibitem[Tevet et~al\mbox{.}(2022)]%
        {tevet2022motionclip}
\bibfield{author}{\bibinfo{person}{Guy Tevet}, \bibinfo{person}{Brian Gordon},
  \bibinfo{person}{Amir Hertz}, \bibinfo{person}{Amit~H Bermano}, {and}
  \bibinfo{person}{Daniel Cohen-Or}.} \bibinfo{year}{2022}\natexlab{}.
\newblock \showarticletitle{Motionclip: Exposing human motion generation to
  clip space}. In \bibinfo{booktitle}{\emph{ECCV}}. Springer,
  \bibinfo{pages}{358--374}.
\newblock


\bibitem[Thies et~al\mbox{.}(2019)]%
        {thies2019deferred}
\bibfield{author}{\bibinfo{person}{Justus Thies}, \bibinfo{person}{Michael
  Zollh{\"o}fer}, {and} \bibinfo{person}{Matthias Nie{\ss}ner}.}
  \bibinfo{year}{2019}\natexlab{}.
\newblock \showarticletitle{Deferred neural rendering: Image synthesis using
  neural textures}.
\newblock \bibinfo{journal}{\emph{Acm Transactions on Graphics (TOG)}}
  \bibinfo{volume}{38}, \bibinfo{number}{4} (\bibinfo{year}{2019}),
  \bibinfo{pages}{1--12}.
\newblock


\bibitem[Tiwari et~al\mbox{.}(2022)]%
        {tiwari2022pose}
\bibfield{author}{\bibinfo{person}{Garvita Tiwari}, \bibinfo{person}{Dimitrije
  Anti{\'c}}, \bibinfo{person}{Jan~Eric Lenssen}, \bibinfo{person}{Nikolaos
  Sarafianos}, \bibinfo{person}{Tony Tung}, {and} \bibinfo{person}{Gerard
  Pons-Moll}.} \bibinfo{year}{2022}\natexlab{}.
\newblock \showarticletitle{Pose-ndf: Modeling human pose manifolds with neural
  distance fields}. In \bibinfo{booktitle}{\emph{ECCV}}. Springer,
  \bibinfo{pages}{572--589}.
\newblock


\bibitem[Tiwari et~al\mbox{.}(2021)]%
        {tiwari2021neural}
\bibfield{author}{\bibinfo{person}{Garvita Tiwari}, \bibinfo{person}{Nikolaos
  Sarafianos}, \bibinfo{person}{Tony Tung}, {and} \bibinfo{person}{Gerard
  Pons-Moll}.} \bibinfo{year}{2021}\natexlab{}.
\newblock \showarticletitle{Neural-GIF: Neural generalized implicit functions
  for animating people in clothing}. In \bibinfo{booktitle}{\emph{ICCV}}.
  \bibinfo{pages}{11708--11718}.
\newblock


\bibitem[Wang et~al\mbox{.}(2021)]%
        {wang2021metaavatar}
\bibfield{author}{\bibinfo{person}{Shaofei Wang}, \bibinfo{person}{Marko
  Mihajlovic}, \bibinfo{person}{Qianli Ma}, \bibinfo{person}{Andreas Geiger},
  {and} \bibinfo{person}{Siyu Tang}.} \bibinfo{year}{2021}\natexlab{}.
\newblock \showarticletitle{Metaavatar: Learning animatable clothed human
  models from few depth images}.
\newblock \bibinfo{journal}{\emph{NeurIPS}}  \bibinfo{volume}{34}
  (\bibinfo{year}{2021}).
\newblock


\bibitem[Wang et~al\mbox{.}(2022)]%
        {wang2022arah}
\bibfield{author}{\bibinfo{person}{Shaofei Wang}, \bibinfo{person}{Katja
  Schwarz}, \bibinfo{person}{Andreas Geiger}, {and} \bibinfo{person}{Siyu
  Tang}.} \bibinfo{year}{2022}\natexlab{}.
\newblock \showarticletitle{Arah: Animatable volume rendering of articulated
  human sdfs}. In \bibinfo{booktitle}{\emph{ECCV}}. Springer,
  \bibinfo{pages}{1--19}.
\newblock


\bibitem[Wang et~al\mbox{.}(2004)]%
        {wang2004image}
\bibfield{author}{\bibinfo{person}{Zhou Wang}, \bibinfo{person}{Alan~C Bovik},
  \bibinfo{person}{Hamid~R Sheikh}, {and} \bibinfo{person}{Eero~P Simoncelli}.}
  \bibinfo{year}{2004}\natexlab{}.
\newblock \showarticletitle{Image quality assessment: from error visibility to
  structural similarity}.
\newblock \bibinfo{journal}{\emph{IEEE transactions on image processing}}
  \bibinfo{volume}{13}, \bibinfo{number}{4} (\bibinfo{year}{2004}),
  \bibinfo{pages}{600--612}.
\newblock


\bibitem[Weng et~al\mbox{.}(2022)]%
        {weng2022humannerf}
\bibfield{author}{\bibinfo{person}{Chung-Yi Weng}, \bibinfo{person}{Brian
  Curless}, \bibinfo{person}{Pratul~P Srinivasan}, \bibinfo{person}{Jonathan~T
  Barron}, {and} \bibinfo{person}{Ira Kemelmacher-Shlizerman}.}
  \bibinfo{year}{2022}\natexlab{}.
\newblock \showarticletitle{Humannerf: Free-viewpoint rendering of moving
  people from monocular video}. In \bibinfo{booktitle}{\emph{CVPR}}.
  \bibinfo{pages}{16210--16220}.
\newblock


\bibitem[Xiang et~al\mbox{.}(2022)]%
        {xiang2022dressing}
\bibfield{author}{\bibinfo{person}{Donglai Xiang}, \bibinfo{person}{Timur
  Bagautdinov}, \bibinfo{person}{Tuur Stuyck}, \bibinfo{person}{Fabian Prada},
  \bibinfo{person}{Javier Romero}, \bibinfo{person}{Weipeng Xu},
  \bibinfo{person}{Shunsuke Saito}, \bibinfo{person}{Jingfan Guo},
  \bibinfo{person}{Breannan Smith}, \bibinfo{person}{Takaaki Shiratori},
  {et~al\mbox{.}}} \bibinfo{year}{2022}\natexlab{}.
\newblock \showarticletitle{Dressing Avatars: Deep Photorealistic Appearance
  for Physically Simulated Clothing}.
\newblock \bibinfo{journal}{\emph{ACM Transactions on Graphics (TOG)}}
  \bibinfo{volume}{41}, \bibinfo{number}{6} (\bibinfo{year}{2022}),
  \bibinfo{pages}{1--15}.
\newblock


\bibitem[Xiang et~al\mbox{.}(2021)]%
        {xiang2021modeling}
\bibfield{author}{\bibinfo{person}{Donglai Xiang}, \bibinfo{person}{Fabian
  Prada}, \bibinfo{person}{Timur Bagautdinov}, \bibinfo{person}{Weipeng Xu},
  \bibinfo{person}{Yuan Dong}, \bibinfo{person}{He Wen},
  \bibinfo{person}{Jessica Hodgins}, {and} \bibinfo{person}{Chenglei Wu}.}
  \bibinfo{year}{2021}\natexlab{}.
\newblock \showarticletitle{Modeling clothing as a separate layer for an
  animatable human avatar}.
\newblock \bibinfo{journal}{\emph{TOG}} \bibinfo{volume}{40},
  \bibinfo{number}{6} (\bibinfo{year}{2021}), \bibinfo{pages}{1--15}.
\newblock


\bibitem[Xu et~al\mbox{.}(2011)]%
        {xu2011video}
\bibfield{author}{\bibinfo{person}{Feng Xu}, \bibinfo{person}{Yebin Liu},
  \bibinfo{person}{Carsten Stoll}, \bibinfo{person}{James Tompkin},
  \bibinfo{person}{Gaurav Bharaj}, \bibinfo{person}{Qionghai Dai},
  \bibinfo{person}{Hans-Peter Seidel}, \bibinfo{person}{Jan Kautz}, {and}
  \bibinfo{person}{Christian Theobalt}.} \bibinfo{year}{2011}\natexlab{}.
\newblock \showarticletitle{Video-based characters: creating new human
  performances from a multi-view video database}.
\newblock \bibinfo{journal}{\emph{TOG}} \bibinfo{volume}{30},
  \bibinfo{number}{4} (\bibinfo{year}{2011}), \bibinfo{pages}{1--10}.
\newblock


\bibitem[Yariv et~al\mbox{.}(2021)]%
        {yariv2021volume}
\bibfield{author}{\bibinfo{person}{Lior Yariv}, \bibinfo{person}{Jiatao Gu},
  \bibinfo{person}{Yoni Kasten}, {and} \bibinfo{person}{Yaron Lipman}.}
  \bibinfo{year}{2021}\natexlab{}.
\newblock \showarticletitle{Volume rendering of neural implicit surfaces}.
\newblock \bibinfo{journal}{\emph{NeurIPS}}  \bibinfo{volume}{34}
  (\bibinfo{year}{2021}), \bibinfo{pages}{4805--4815}.
\newblock


\bibitem[Yoon et~al\mbox{.}(2022)]%
        {yoon2022learning}
\bibfield{author}{\bibinfo{person}{Jae~Shin Yoon}, \bibinfo{person}{Duygu
  Ceylan}, \bibinfo{person}{Tuanfeng~Y Wang}, \bibinfo{person}{Jingwan Lu},
  \bibinfo{person}{Jimei Yang}, \bibinfo{person}{Zhixin Shu}, {and}
  \bibinfo{person}{Hyun~Soo Park}.} \bibinfo{year}{2022}\natexlab{}.
\newblock \showarticletitle{Learning motion-dependent appearance for
  high-fidelity rendering of dynamic humans from a single camera}. In
  \bibinfo{booktitle}{\emph{CVPR}}. \bibinfo{pages}{3407--3417}.
\newblock


\bibitem[Yu et~al\mbox{.}(2021)]%
        {yu2021plenoctrees}
\bibfield{author}{\bibinfo{person}{Alex Yu}, \bibinfo{person}{Ruilong Li},
  \bibinfo{person}{Matthew Tancik}, \bibinfo{person}{Hao Li},
  \bibinfo{person}{Ren Ng}, {and} \bibinfo{person}{Angjoo Kanazawa}.}
  \bibinfo{year}{2021}\natexlab{}.
\newblock \showarticletitle{Plenoctrees for real-time rendering of neural
  radiance fields}. In \bibinfo{booktitle}{\emph{ICCV}}.
  \bibinfo{pages}{5752--5761}.
\newblock


\bibitem[Zhang et~al\mbox{.}(2023)]%
        {zhang2023closet}
\bibfield{author}{\bibinfo{person}{Hongwen Zhang}, \bibinfo{person}{Siyou Lin},
  \bibinfo{person}{Ruizhi Shao}, \bibinfo{person}{Yuxiang Zhang},
  \bibinfo{person}{Zerong Zheng}, \bibinfo{person}{Han Huang},
  \bibinfo{person}{Yandong Guo}, {and} \bibinfo{person}{Yebin Liu}.}
  \bibinfo{year}{2023}\natexlab{}.
\newblock \showarticletitle{CloSET: Modeling Clothed Humans on Continuous
  Surface with Explicit Template Decomposition}. In
  \bibinfo{booktitle}{\emph{CVPR}}.
\newblock


\bibitem[Zhang et~al\mbox{.}(2018)]%
        {zhang2018unreasonable}
\bibfield{author}{\bibinfo{person}{Richard Zhang}, \bibinfo{person}{Phillip
  Isola}, \bibinfo{person}{Alexei~A Efros}, \bibinfo{person}{Eli Shechtman},
  {and} \bibinfo{person}{Oliver Wang}.} \bibinfo{year}{2018}\natexlab{}.
\newblock \showarticletitle{The unreasonable effectiveness of deep features as
  a perceptual metric}. In \bibinfo{booktitle}{\emph{CVPR}}.
  \bibinfo{pages}{586--595}.
\newblock


\bibitem[Zheng et~al\mbox{.}(2022)]%
        {zheng2022structured}
\bibfield{author}{\bibinfo{person}{Zerong Zheng}, \bibinfo{person}{Han Huang},
  \bibinfo{person}{Tao Yu}, \bibinfo{person}{Hongwen Zhang},
  \bibinfo{person}{Yandong Guo}, {and} \bibinfo{person}{Yebin Liu}.}
  \bibinfo{year}{2022}\natexlab{}.
\newblock \showarticletitle{Structured local radiance fields for human avatar
  modeling}. In \bibinfo{booktitle}{\emph{CVPR}}.
  \bibinfo{pages}{15893--15903}.
\newblock


\bibitem[Zheng et~al\mbox{.}(2023)]%
        {zheng2023avatar}
\bibfield{author}{\bibinfo{person}{Zerong Zheng}, \bibinfo{person}{Xiaochen
  Zhao}, \bibinfo{person}{Hongwen Zhang}, \bibinfo{person}{Boning Liu}, {and}
  \bibinfo{person}{Yebin Liu}.} \bibinfo{year}{2023}\natexlab{}.
\newblock \showarticletitle{AvatarReX: Real-time Expressive Full-body Avatars}.
\newblock \bibinfo{journal}{\emph{ACM Transactions on Graphics (TOG)}}
  \bibinfo{volume}{42}, \bibinfo{number}{4} (\bibinfo{year}{2023}).
\newblock
\urldef\tempurl%
\url{https://doi.org/10.1145/3592101}
\showDOI{\tempurl}


\bibitem[Zhou et~al\mbox{.}(2019)]%
        {zhou2019continuity}
\bibfield{author}{\bibinfo{person}{Yi Zhou}, \bibinfo{person}{Connelly Barnes},
  \bibinfo{person}{Jingwan Lu}, \bibinfo{person}{Jimei Yang}, {and}
  \bibinfo{person}{Hao Li}.} \bibinfo{year}{2019}\natexlab{}.
\newblock \showarticletitle{On the continuity of rotation representations in
  neural networks}. In \bibinfo{booktitle}{\emph{CVPR}}.
  \bibinfo{pages}{5745--5753}.
\newblock


\end{thebibliography}
